\documentclass[conference]{IEEEtran}
\IEEEoverridecommandlockouts
\usepackage{cite}
\usepackage{amsmath,amssymb,amsfonts}
\usepackage{algorithmic}
\usepackage{graphicx}
\usepackage{textcomp}
\usepackage{subfigure}
\usepackage{multirow}
\usepackage[table]{xcolor}
\usepackage{booktabs}

\newtheorem{theorem}{Theorem}
\usepackage{algorithm}
\def\BibTeX{{\rm B\kern-.05em{\sc i\kern-.025em b}\kern-.08em
    T\kern-.1667em\lower.7ex\hbox{E}\kern-.125emX}}
\begin{document}

\title{Robust Deep Reinforcement Learning with Adaptive\\ Adversarial Perturbations in Action Space}

\DeclareRobustCommand*{\IEEEauthorrefmark}[1]{%
    \raisebox{0pt}[0pt][0pt]{\textsuperscript{\footnotesize\ensuremath{#1}}}}
\author{
   \IEEEauthorblockN{\textbf{Qianmei Liu\IEEEauthorrefmark{1}, Yufei Kuang\IEEEauthorrefmark{1}, Jie Wang\IEEEauthorrefmark{1,2*}}}
   \IEEEauthorblockA{\IEEEauthorrefmark{1}University of Science and Technology of China}
   \IEEEauthorblockA{\IEEEauthorrefmark{2}Institute of Artificial Intelligence, Hefei Comprehensive National Science Center}
   \IEEEauthorblockA{\{liuqianmei, yfkuang\}@mail.ustc.edu.cn}
   \IEEEauthorblockA{{jiewangx}@ustc.edu.cn}
   \thanks{*Corresponding author.}
}

\maketitle

\begin{abstract}
Deep reinforcement learning (DRL) algorithms can suffer from modeling errors between the simulation and the real world. Many studies use adversarial learning to generate perturbation during training process to model the discrepancy and improve the robustness of DRL. However, most of these approaches use a fixed parameter to control the intensity of the adversarial perturbation, which can lead to a trade-off between average performance and robustness. In fact, finding the optimal parameter of the perturbation is challenging, as excessive perturbations may destabilize training and compromise agent performance, while insufficient perturbations may not impart enough information to enhance robustness. To keep the training stable while improving robustness, we propose a simple but effective method, namely, Adaptive Adversarial Perturbation (A2P), which can dynamically select appropriate adversarial perturbations for each sample. Specifically, we propose an adaptive adversarial coefficient framework to adjust the effect of the adversarial perturbation during training. By designing a metric for the current intensity of the perturbation, our method can calculate the suitable perturbation levels based on the current relative performance. The appealing feature of our method is that it is simple to deploy in real-world applications and does not require accessing the simulator in advance. The experiments in MuJoCo show that our method can improve the training stability and learn a robust policy when migrated to different test environments. The code is available
at https://github.com/Lqm00/A2P-SAC.
\end{abstract}


\section{Introduction}
Deep reinforcement learning algorithms have recently made significant strides in fields such as gaming\cite{mnih2015human, lillicrap2015continuous,vinyals2019grandmaster}, robot control\cite{levine2016end,andrychowicz2020learning,sangiovanni2018deep}, and autonomous driving\cite{sallab2017deep,aradi2020survey}. However, most of these tasks are trained in simulation, which may not match the real-world dynamics. This mismatch can cause the policies to perform poorly or even fail when transferred to the real world\cite{jiang2021monotonic,tzeng2015towards,peng2018sim}. This can have serious consequences for robotics and safety-critical systems. Therefore, it is essential to ensure the reliability and robustness of the agent before applying DRL algorithms in practice. To achieve this goal, we need to learn policies that can handle environmental perturbations, sensor errors, and control uncertainties.

Many recent studies have applied adversarial training to the DRL algorithm to solve the generalization issue of deep neural networks. It is inspired by the success of adversarial training in enhancing the generalization for supervised learning tasks such as image classification. One simple approach is based on robust Markov decision process (RMDP)\cite{iyengar2005robust} framework, which trains on an ensemble of worst-case transition dynamics. For example, Ensemble Policy Optimization (EPOpt)\cite{rajeswaran2016epopt} algorithm samples and weights model instances where the policy performs poorly in the source distribution. However, this approach requires defining uncertainty sets, which can be difficult. Another approach is to add perturbations to the system during training to generate worst-case trajectories. For example, RARL\cite{pinto2017robust} models the perturbation on the action space, while SA-MDP\cite{zhang2020robust} models the perturbation on the state observations. The objective of adversarial training is to learn a robust policy that maximizes the cumulative reward while defending against adversarial attacks. Furthermore, we observe that the intensity of the perturbation, which is a parameter that reflects the degree of deviation between the adversarial sample and the original sample, also has a significant impact on the robustness.

Despite the success of the current approach, determining the intensity of the perturbation is still challenging. The intensity of the perturbation affects the trade-off between the robustness and the average performance. If the perturbation is too strong, it can cause biased sampling and unstable training, which can degrade the policy performance. If the perturbation is too weak, it can fail to improve the robustness of the policy in target environments. Moreover, current approaches rely on manual tuning of the perturbation intensity, which can be tedious and inefficient. For example, RARL\cite{pinto2017robust} uses a fixed adversarial force for each task, while NR-MDP\cite{tessler2019action} manually adjusts the adversarial perturbation coefficient. However, it is hard to find a suitable value that balances robustness and average performance.

Motivated by these ideas, we propose a novel method, namely, Adaptive Adversarial Perturbation(A2P), aiming to train a robust policy with appropriate levels. First, an adjusting adversarial coefficient framework is proposed to  dynamically control the effect of the adversarial behavior in the training process. While it is intuitively reasonable to assign higher perturbations to examples with smaller step size, these methods may suffer from model bias and misled to hurt model performance. Second, we design a  metric of the current perturbation intensity which allow the model to calculate the appropriate perturbations levels based on the current relative performance. Finally, as similar actions may lead to similar results in one aspect, our method is applied to a classic experiment that adds perturbations to the action space. In this way, we can sample worst-case trajectories while also improving the robustness of the policy. 

To summarize, our contributions in this paper are:
\begin{itemize}
\item {In this paper, we study a prevalent trade-off in robust adversarial reinforcement learning: how to achieve both high average performance and robustness for the model under adversarial training. We conduct experiments to examine how the model performance varies with different levels of adversarial perturbations.}
\item {We propose an novel method A2P that can dynamically adjust the adversarial perturbation intensity according to the training situation to alleviate the trade-off. The strength of our method lies in its simplicity, making it ideal for real-world applications without necessitating pre-existing access to or modeling of the simulator.}
	\item {We conducted extensive experiments on MuJoCo tasks based on the SAC algorithm as a baseline. The results show that A2P performs better than baseline when the environmental parameters change, which demonstrates that our method learns a more robust policy.}
\end{itemize}

\section{Related Work}
\subsection{Robust Reinforcement Learning (RRL)}
RRL aims to learn a robust optimal policy that considers model uncertainty in transition dynamics to systematically mitigate the sensitivity of the policy in a perturbed environment. Various of algorithms have presented to search the robust optimal policy with model uncertainty during the learning process. For instance, Domain Randomization (DR) generates environments with different transition dynamics \cite{tobin2017domain}. Others use Wasserstein distance to define the set of uncertainties in the dynamics \cite{yang2017convex}. Robust MPO (RMPO) learns a robust policy using the transition dynamical set that models in advance \cite{mankowitz2019robust}. Wasserstein Robust Reinforcement Learning (WR2L) proposes to co-train environmental parameters with the protagonist \cite{abdullah2019wasserstein}. While these methods seem to be effective in practice, they require simulator access and control. In contrast, our approach is easier to deploy and does not require additional knowledge of perturbations and a specially designed simulator.

\subsection{Adversarial Training (AT)}
AT is a training method that introduces noise to regularize parameters and improve model robustness and generalization. In the field of computer vision, classical approaches \cite{kour2014real,madry2017towards,athalye2018obfuscated} use loss gradients relative to the input image to generate adversarial samples that are mixed with the original samples for training. In DRL, there are three main ways to generate adversarial samples. First, it generates the adversarial samples by using loss gradients as the same as the computer vision. For example, an method used active computation of physically plausible adversarial examples that are generated by a weak FGSM \cite{goodfellow2014explaining} based attack with policy gradient to train the agent \cite{mandlekar2017adversarially}. Second, many methods use adversarial boundary sets to select perturbation. For example, SA-MDP proposed a method to add noise that selects from boundary sets on the state observation space to simulate natural measurement error \cite{zhang2020robust}. Compared to this method, SCPO \cite{kuang2022learning} adds perturbations on state space to convert real environmental perturbations that are difficult to model in advance. Finally, many studies have proposed a method that uses adversarial policy to generate perturbation. RARL incorporates an adversarial agent that applies additional forces to the protagonist \cite{pinto2017robust}. Other study proposed two methods to add noise on the action space \cite{tessler2019action}. One is to select the adversarial action with a certain probability, and the other is that an adversary adds perturbations to the selected action. Therefore, based on the fact that similar behaviors always lead to similar results, we propose a method to add perturbations directly on the action space.

\section{Preliminaries}
In this section, we briefly introduce the preliminaries.
\subsection{Markov Decision Process (MDP)}
In reinforcement learning, the interaction between the agent and the environment is often modeled through a Markov decision process (MDP). The process can be represented by the tuple $(\mathcal{S},\mathcal{A},\mathcal{P},\mathcal{R},\gamma)$. Here $\mathcal{S}$ is  the state space, $\mathcal{A}$ is the action space, and $\mathcal{P}:\mathcal{S}\times \mathcal{A}\rightarrow \Delta(\mathcal{S})$ is the transition dynamic, where $\Delta(\mathcal{S})$ denotes the probability distribution on $\mathcal{S}$. $R:\mathcal{S}\times \mathcal{A}\rightarrow \Delta(\mathcal{R})$ is the reward function, where $\Delta(\mathcal{R})$ denotes the probability distribution on $\mathcal{R}$. However, in many reinforcement learning problems, the reward signal is often an artificially designed deterministic function, so researchers can also define $\mathcal{R}$ as $\mathcal{S}\times \mathcal{A}\rightarrow[0,r_{max}]$. For example, $\mathcal{R}(s,a)$ denotes the reward obtained after taking action $a$ in state $s$. $\gamma\in(0,1)$ is the discount factor. Our approach is based on adding disturbances to the MDP under continuous action space, in order to simulate the differences between transition dynamics.
\begin{figure*}[ht]
    \centering
    \subfigcapskip=-5pt
\subfigure[ ]{
\label{Fig.sub.1}
\includegraphics[width=0.3\linewidth]{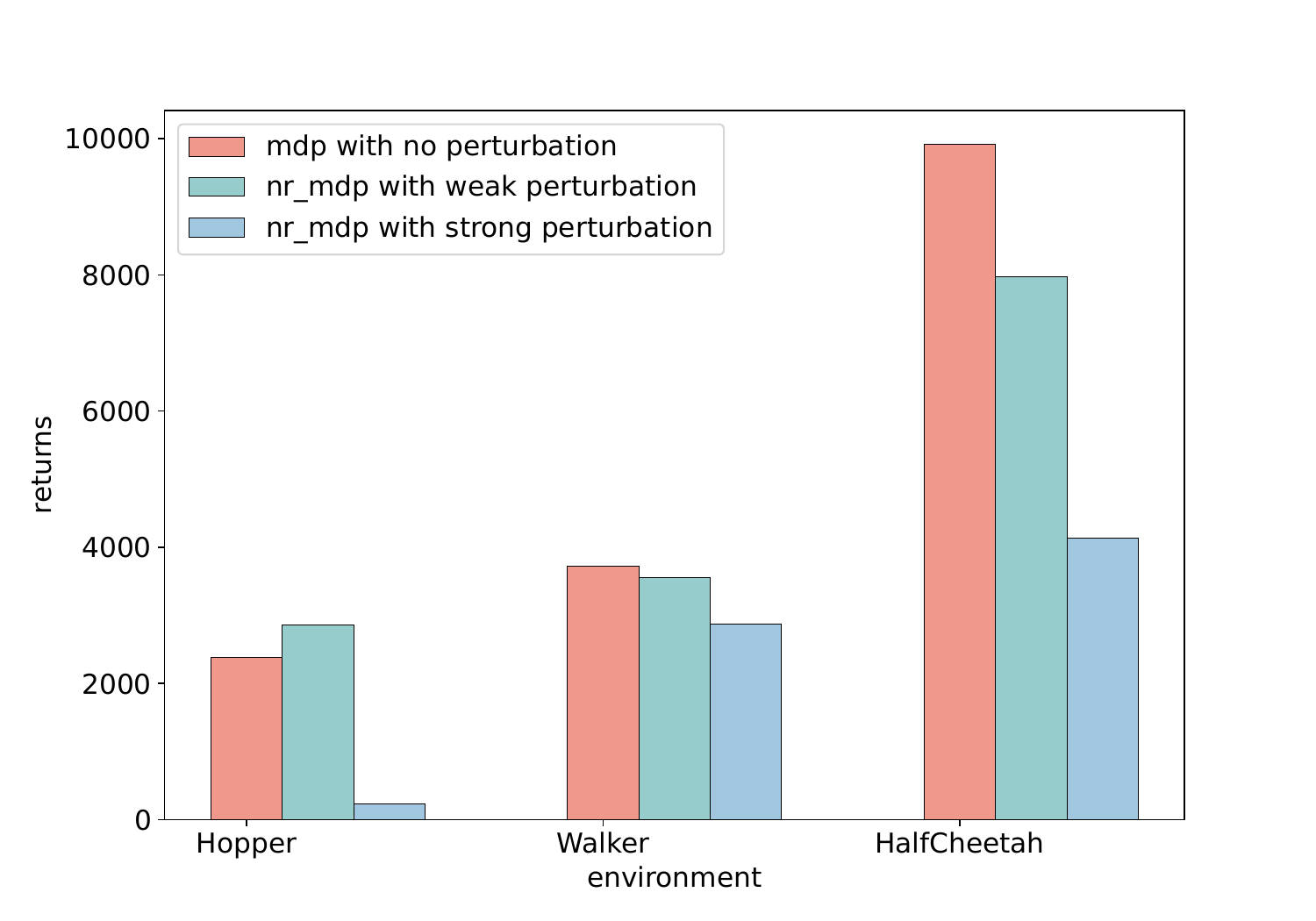}}
\hspace{-3mm}
\subfigure[ ]{
\label{Fig.sub.2}
\includegraphics[width=0.3\linewidth]{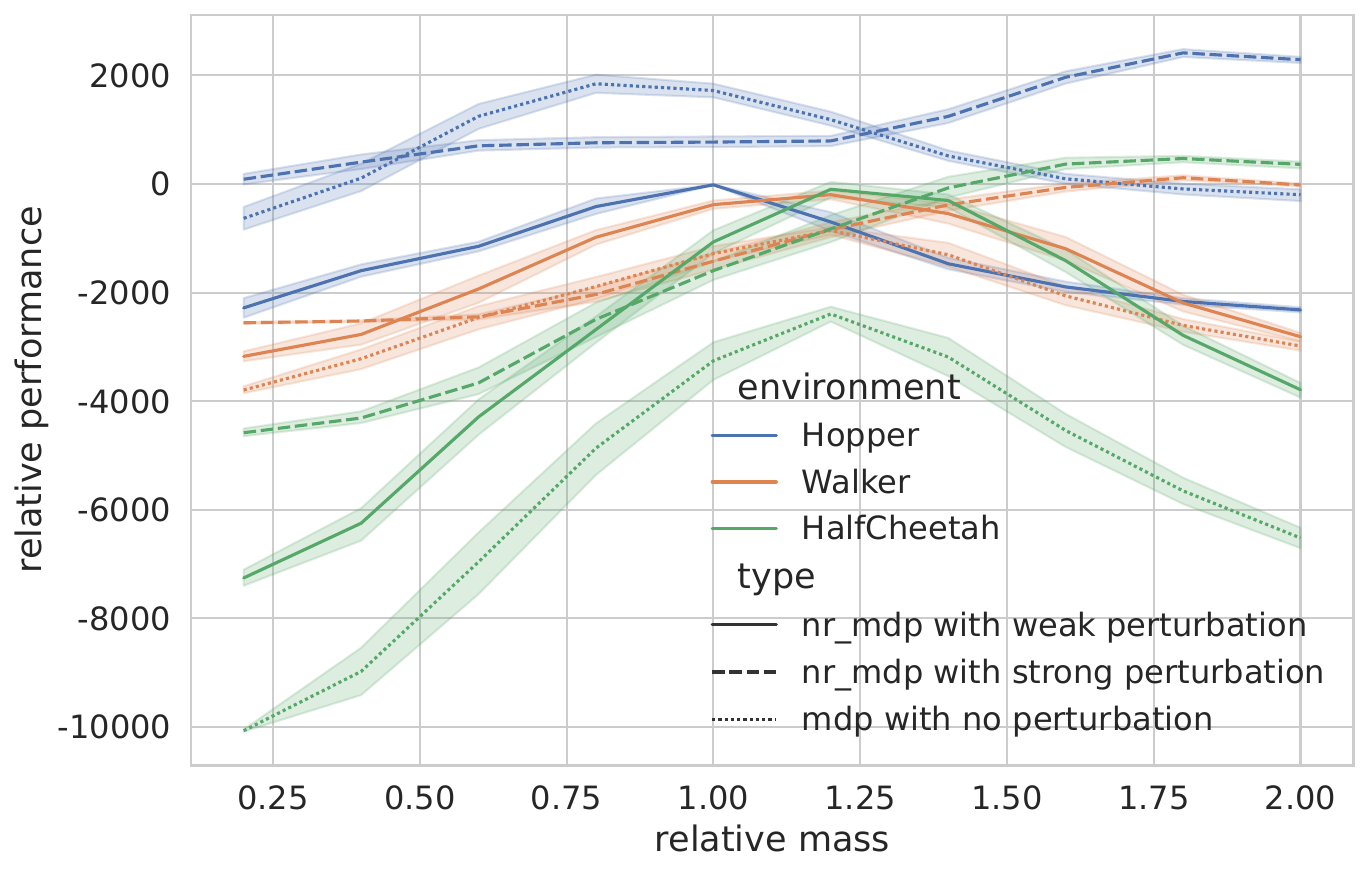}}
\hspace{-2mm}
\subfigure[ ]{
\label{Fig.sub.3}
\includegraphics[width=0.3\linewidth]{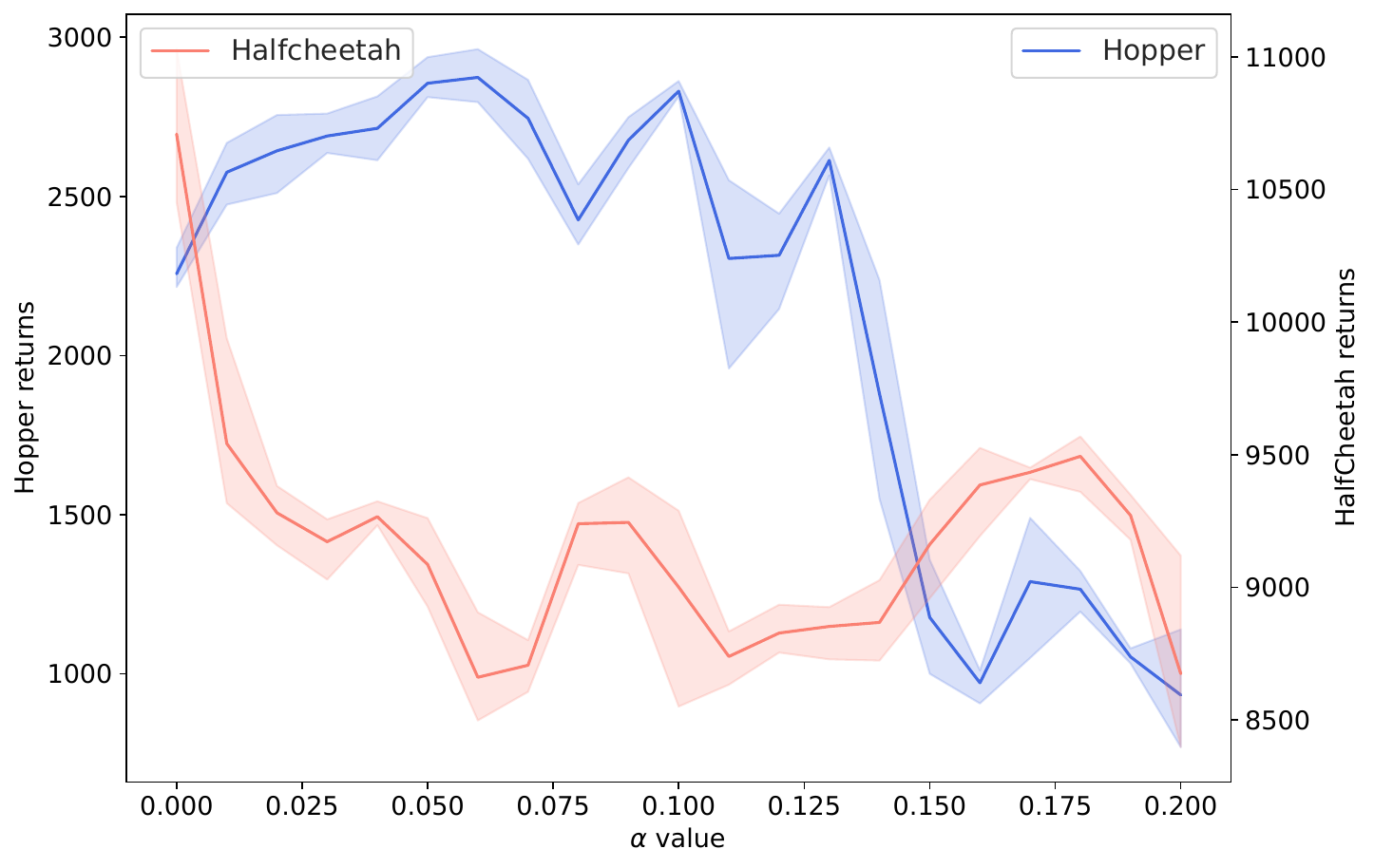}}
\vspace{-2mm}
\caption{(a) Comparison of performance in source environments with different perturbation methods. (b) Comparison of the relative performance, which compares with mass = 1.0, with different perturbation methods in testing environments. (c) Effects of range-specific perturbations on model performance in the source environment.}
\label{fig:motivation}
\vspace{-5mm}
\end{figure*}
\subsection{Zero-sum Game}
Zero-sum game means that the opponents involved in this game process, the benefit of one side, it must mean the loss of the other side, so the gains and losses of both sides of the game must add up to zero. The benefit of the whole process is not increased. This process is usually represented by the tuple $(\mathcal{S},\mathcal{A},\mathcal{O},\mathcal{P},\mathcal{R},\mathcal{\gamma})$, where $O$ is other palyer's action set. In adversarial reinforcement learning, the adversary and the protagonist will reach an equilibrium in which the adversary minimizes the reward function and the protagonist maximizes the reward function by training in a game-based learning approach. The objective is defined as
\begin{equation}
    \begin{aligned}
        V^\ast(s)=&\max\limits_{\pi(s)\in\Pi}\min\limits_{o\in \mathcal{O}}\sum\limits_{a\in\mathcal{A}}\pi(a\mid s)[\mathcal{R}(s,a,o)\\&+\gamma\sum\limits_{s^\prime\in\mathcal{S}}\mathcal{P}(s^\prime\mid\,s,a,o)V^\ast(s^\prime)].
    \end{aligned}
\end{equation}

\subsection{Robustness and Average Performance} Robust Markov decision process (RMDP) aims to maximize the cumulative reward of the worst trajectory under the transition dynamic. The environment is selected from a set of transition dynamics that minimize the reward. In the RMDP, a policy $\pi$ is robust for all the worst-case trajectories that are chosen from the $p(s^\prime\mid\,s,a)$, and we have
\begin{equation}
\pi=\mathop{\arg\max}\limits_{\pi\in\Pi}\min\limits_{p(s^\prime\mid\,s,a)} (r(s,a)+\gamma\,Q(s^\prime,a^\prime)).
\end{equation} Also, we expect $\pi$ to maximize the return under the $p(\cdot\mid\,s,a)$ of the source environment in RRL. The goal of our method is to find a policy with the highest performance. It's worth mentioning that many works usually use optimal performance to refer to standard test performance without perturbation, while we use the average performance with the adversarial perturbation in our paper.

\section{Method}
\subsection{Motivation}
In the reinforcement learning based on adversarial training, the intensity of the perturbation is determined by a parameter. For example, the noise is sampled from a ball of radius $\epsilon$ in the parameter space, and the size of the noise is also determined by $\epsilon$. We can refer to $\epsilon$ as the adversarial coefficient.

Intuitively, we guess that the intensity of the adversarial perturbation affects the effectiveness of adversarial training. First, excessive perturbation may destabilize training, while it may improve the performance in more challenging environments. Second, insufficient perturbation may not impart enough information to enhance robustness when it is transferred to the test environment where the environmental parameters have changed, which means that the goal of adversarial training is not achieved. Next, we use experiments to illustrate this problem and motivate our proposed method.

We use a simple NR-MDP~\cite{tessler2019action} experiment to demonstrate how perturbation affects adversarial training. Here the adversarial coefficient is defined as $\alpha$. We run adversarial training with $\alpha=0.2$ as a strong perturbation case and $\alpha=0.01$ as a weak one. In Figure~\ref{Fig.sub.1}, we compare the average performance of the different perturbed algorithms and the unperturbed algorithm for the three MuJoCo tasks. We test the average returns of the trained model in the training environment. The results show that the average performance of some tasks, such as Hopper-v3 and HalfCheetah-v2, deteriorates significantly worse under the effect of strong perturbation, while the weakly perturbed method can maintain the unperturbed performance. In Figure~\ref{Fig.sub.2}, we show the relative performance of different perturbed algorithms and the unperturbed algorithm in testing environments that are deployed by changing certain environmental parameters (e.g., mass) of the training environment to simulate errors from real-world tasks. The performance of most environments improves under strong perturbation, which indicates that the strong perturbation algorithm has more potential to learn a robust policy than the weak one. This motivates the following problem:

The goal of RL is to learn a policy that performs well in both the training and testing environments. However, there is often a trade-off between the average performance and the robustness of the policy. A robust policy can withstand strong perturbation, but it may sacrifice some average performance. A non-robust policy can achieve high average performance, but it may fail in extreme cases.

Furthermore, we investigate how the average performance of different tasks varies with different adversarial perturbation intensities. Based on previous experimental studies, we chose $\alpha\in(0,0.2)$ as the range of perturbation intensities. Figure~\ref{Fig.sub.3} shows that the average performance generally deteriorates as $\alpha$ increases, but not in a strictly monotonic manner. Therefore, it is challenging to find an optimal parameter that maximizes the average performance of adversarial training. Moreover, fixed-parameter methods require more effort to search for optimal parameters and need to train multiple times for each task. In contrast, our method only needs to train once for each task and achieves a balance between average performance and robustness.

We confirmed our hypothesis that the intensity of the adversarial perturbation affects the effectiveness of adversarial training, which is evidenced by the trade-off between average performance and robustness. 

\subsection{Adaptive Adversarial Perturbation Markov Decision Process (A2P-MDP)}
From our analysis in motivation, we propose A2P method to dynamically adjust the intensity of perturbations. In this section, we propose the new objective of A2P method.

The robust Markov decision process(RMDP) takes into account disturbances in the transition dynamics at each step and attempts to maximize the cumulative reward of the worst trajectory. The objective of RMDP is defined as 
\begin{equation}
    J_{\mathcal{P}}(\pi)\stackrel {\Delta} {\mathop {=}}\inf\limits_{p\in\mathcal{P}}\mathbf{E}_{p,\pi}[\sum^{\infty}_{t=0}\gamma^{t}r(s_t,a_t)].
\end{equation}
RMDP requires modeling and access to the environment, but in fact real transition dynamic functions are not available for many tasks. Thus, our approach to ensemble modeling of $\mathcal{P}$ can be reduced to the perturbation in $\mathcal{A}$. Next, we formalize the objective function for adaptive perturbation in action space, distinguishing it from the RMDP. The object of our method is defined as
\begin{equation}
\begin{aligned}
J_{\epsilon-\mathcal{A}}^{apdt}(\pi)\stackrel {\Delta} {\mathop {=}}&\,\mathbf{E}_{s_0\sim\,d_0}[\min\limits_{\bar{a}_0\sim\bar{\pi}(\cdot\mid\,s_0)}\mathbf{E}_{a_0\sim\pi(\cdot\mid\,s_0)}[r^{adpt}_{\epsilon-\mathcal{A}}(s_0,a_0,\bar{a}_0)\\&+\gamma\mathbf{E}_{s_1\sim\,p_{\epsilon-\mathcal{A}}^{adpt}(\cdot\mid\,s_0,a_0,\bar{a}_0)}[\min\limits_{\bar{a}_1\sim\bar{\pi}(\cdot\mid\,s_1)}\mathbf{E}_{a_1\sim\pi(\cdot\mid\,s_1)}\\&[r_{\epsilon-\mathcal{A}}^{adpt}(s_1,a_1,\bar{a}_1)+...]],
\end{aligned}
\end{equation}
where $d_o$ represents the initial state on the source environment, action $a$ is selected by the target policy, and action $\bar{a}$ is selected by the adversarial policy. We aim to maximize the objective (4) and call it as adaptive action perturbation Markov decision process (A2P-MDP). The induced reward function and dynamics from A2P-MDP are defined as
\begin{equation}
    r^{apdt}_{\epsilon-A}(s,a,\bar{a})\stackrel {\Delta} {\mathop {=}}\,r(s,(1-\epsilon^{adpt})a+\epsilon^{adpt}\bar{a}),
\end{equation}
\begin{equation}
    \begin{aligned}
        \mathcal{P}^{adpt}_{\epsilon-A}(s^{\prime}\mid\,s,a,\bar{a})\stackrel {\Delta} {\mathop {=}}\,&P(s^{\prime}\mid\,s,(1-\epsilon^{adpt})a+\epsilon^{adpt}\bar{a}),
    \end{aligned}
\end{equation}
where $\epsilon^{adpt}$ is an adaptive parameter that changes dynamically with the training process. It is worth mentioning that when sets $\epsilon^{adpt}$ to 0, our objective is consistent with the original reinforcement learning objective.
\begin{figure}[ht]
\centering
\vspace{-3mm}
\includegraphics[width=0.49\textwidth]{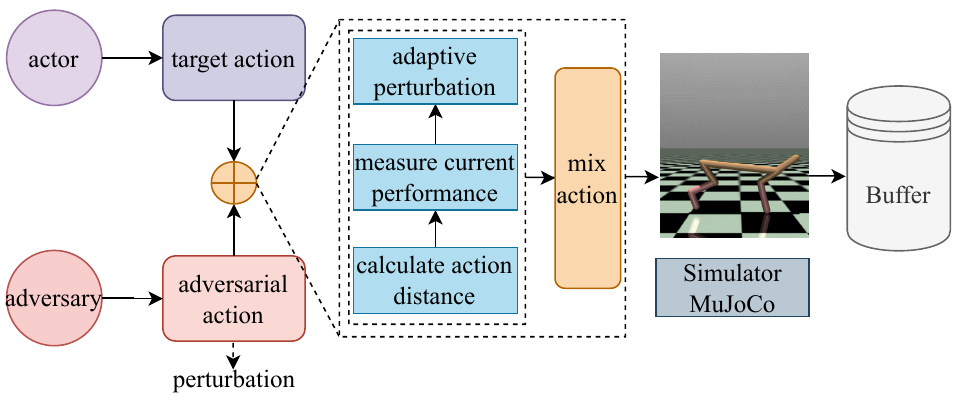}
\vspace{-2mm}
\caption{The general architecture of the A2P method }
\label{fig2:model}
\vspace{-4mm}
\end{figure}
\subsection{Adaptive Adversarial Perturbation in Action Space}
We propose adaptive perturbation to alleviate the trade-off problem during training. The robust policy aims to maximize the cumulative reward and learns optimal actions for the goal. We measure the attack intensity by the distance to the target action from the robust policy. If the adversarial action is close to the target action, we assume it is more likely to be reasonable and weaken the attack accordingly. Thus, we adjust the coefficient $\epsilon$ to control the adversarial action based on the distance. 

First, we need to measure the distance between the target action and the adversarial action. Since we consider continuous control tasks, the actions generated by the two policies under the same state are continuous vectors, and they have the same dimension. To simplify the calculation, we use the Euclidean distance to measure the distance between the action pairs:
\begin{equation}
\begin{aligned}
    d(a,\bar{a})&=\sqrt{\sum_{i=1}^{n}(a[i]-\bar{a}[i])^{2}}\\&
    =\lVert\,a-\bar{a}\rVert^2_2.
\end{aligned}
\end{equation}

Second, we use the moving average of the action distance
\begin{equation}
    d_{j}=\beta\,d_{j-1}+(1-\beta)d
\end{equation}
to adjust the adversarial coefficient $\epsilon_{j}$ for the $a_{j}$ at the j-th epoch. Here, $\beta$ is the momentum factor stabilizing the adversarial coefficient.

Third, We propose a new metric $b$ for evaluating model performance, which is based on the moving average distance and the impact of actions on distance:
\begin{equation}
    b=\mathrm{sgn}(d_{j-1}-d)\sigma(\lvert d_{j-1}-d\rvert).
\end{equation}

Finally, the relationship between the adversarial coefficient and the distance of the actions is as follows: 
\begin{equation}
    \epsilon_{j}=\epsilon_{j-1}+c\cdot\,b,
\end{equation}
where $c$ is a predefined learning rate, and the sign function indicates whether the current perturbation is weaker or stronger than the previous one. We increase or decrease the adversarial coefficient $\epsilon$ accordingly.

The fixed $\epsilon$ has two problems. First, it is hard to choose the best $\epsilon$ for both average performance and robustness. Second, the optimal $\epsilon$ varies for different tasks. We address these issues and propose to trade-off the weights of the two components in the target actions. Here is an adversarial example of our algorithm:
\begin{equation}
    a_{j}^{adpt}=(1-\epsilon_{j})a_{j}+\epsilon_{j}\bar{a}_{j}
\end{equation}
where $a_{j}^{adpt}$ is the final action that interact with the environment with the perturbation. The general architecture of the A2P method is shown in Figure~\ref{fig2:model}.

\subsection{Theoretical Analysis}
In our approach, the perturbation would be very arbitrary if the perturbation coefficient was not controlled. In order to control the adversarial action within a reasonable range, we restrict the perturbation coefficient to [0,1].
Given a fixed policy $\pi$, we can evaluate its performance
(value function). Then, we define the action-value function as
\begin{equation}
    \begin{aligned}
        Q^{adpt}_{\epsilon-\mathcal{A}}(s,a,\bar{a})\mathop{=}&r(s,a,\bar{a})+\gamma\mathbf{E}_{s_{1}\sim{p_0(\cdot\mid{s,a,\bar{a}})}}[\min\limits_{\bar{a}_1\sim\bar{\pi}(\cdot\mid s_1)}\\&\mathbf{E}_{a_1\sim\pi(\cdot\mid s_1)}[Q^{adpt}_{\epsilon-A}(s_1,a_1,\bar{a}_1)]],
    \end{aligned}
    \label{eq7}
\end{equation}
where $Q^{adpt}_{\epsilon-\mathcal{A}}(s,a,\bar{a})\mathop{=}Q(s,(1-\epsilon^{adpt})a+\epsilon^{adpt}\bar{a})$. The above formula shows that the $Q$-value is determined by both the target action and the adversarial action. To simplify the formulation, we define the adaptive action as $a^{adpt}=(1-\epsilon^{adpt})a+\epsilon^{adpt}\bar{a}$. $Q^{adpt}_{\epsilon-\mathcal{A}}$ is the fixed point of the following operator
\begin{equation}
    \begin{aligned}
        \mathbf{B}^{adpt}_{\epsilon-\mathcal{A}}Q(s,a^{adpt})\mathop{=}&r(s,a^{adpt})+\gamma\mathbf{E}_{s_1\sim{p_0(\cdot\mid{s,a^{adpt}})}}[\min\limits_{\bar{a_1}\sim\bar{\pi}(\cdot\mid s_1)}\\&\mathbf{E}_{a_1\sim\pi(\cdot\mid s_1)}[Q(s_1,a^{adpt}_1)]],
    \end{aligned}
    \label{eq8}
\end{equation}
where $\mathbf{B}^{adpt}_{\epsilon-\mathcal{A}}$ is the adaptive adversarial Bellman operator. We consider $\mathbf{B}^{adpt}_{\epsilon-\mathcal{A}}$ to be a contraction mapping of $Q$-value.
\begin{theorem}[Bellman Contraction.]
    For any $\epsilon\in[0,1]$ and any fixed policy, the adaptive adversarial Bellman operator $\mathbf{B}^{adpt}_{\epsilon-\mathcal{A}}$ in Equation \eqref{eq8} a contraction that converges to $Q^{adpt}_{\epsilon-\mathcal{A}}$ in Equation \eqref{eq7}.
    \label{theorem1}
\end{theorem}

Theorem~\ref{theorem1} states that given a fixed policy $\pi$, we can evaluate its performance (value function) at the current optimal perturbation, through a Bellman contraction. Similar to the original MDP, the objective function in A2P-MDP is to maximize the cumulative reward, and it will eventually converge to the optimal value function.
\begin{theorem}[Policy Improvement.]
    For any greedy policy $\pi^{old}$, the greedy policy $\pi^{new}$ obtained using the action-value function $Q^{adpt}_{\epsilon-\mathcal{A}}$ is a policy improvement on $\pi^{old}$.The specific rule of the policy improvement is
\begin{equation}
    \begin{aligned}
        \pi_{new}(\cdot\mid\,s)\stackrel {\Delta} {\mathop {=}}\mathop{\arg\max}\limits_{\pi\in\Pi}\min\limits_{\bar{a}\in\bar{\pi}}\mathbf{E}_{a\in\pi}[Q^{\pi_{old}}_{\epsilon-\mathcal{A}}(s,a^{adpt})]
    \end{aligned}
\end{equation}
Then $Q^{\pi_{new}}_{\epsilon-\mathcal{A}}(s,a^{adpt})\geq\,Q^{\pi_{old}}_{\epsilon-\mathcal{A}}(s,a^{adpt})$ for all $(s,a)\in\mathcal{S}\times\mathcal{A}$.
\label{theorem2}
\end{theorem}

The new policy is updated by maximizing the $Q^{\pi_{old}}$, then the $Q^{\pi_{new}}$ of the action generated by the new policy must be no less than the $Q^{\pi_{old}}$. By setting $\epsilon$ to 0, A2P method recovers the original policy iteration algorithm.
\begin{algorithm}[h]
\caption{Adaptive Adversarial Perturbations Soft Actor-Critic}
\label{alg:algorithm}
\begin{algorithmic}[1]
\STATE \textbf{Input}: Randomly initialize critic network $Q_{\theta_{1}}$, $Q_{\theta_{2}}$, actor $\pi_{\phi}$, adversary $\bar{\pi}_{\bar\phi}$. Initial temperature parameter $\alpha$. Step size $\gamma$. Adaptive coefficient $\epsilon$. $\bar{\theta}_{1}\xleftarrow{}\theta_{1},$ $\bar{\theta}_{2}\xleftarrow{}\theta_{2},$
		$\mathcal{D}\xleftarrow{}\emptyset$.

\STATE Let $d = 0, \epsilon = 0.1$.
\FOR{each environment step}
\STATE calculate the distance of actions\\
\STATE update the moving average distance $d$.\\
\STATE adaptive adversarial coefficient $\epsilon$.\\
\STATE $\mathrm{a}_t=(1-\epsilon) \pi\left(\cdot \mid \mathrm{s}_t\right)+\epsilon\bar{\pi}\left(\cdot \mid \mathrm{s}_t\right).$\\
\STATE $\mathcal{D} \leftarrow \mathcal{D} \cup\left\{\mathrm{s}_t, \mathrm{a}_t, r\left(\mathrm{s}_t, \mathrm{a}_t\right), \mathrm{s}_{t+1}\right\}.$\\
\FOR{each training step}
\STATE  sample batch from replay buffer\\
\STATE $
    \theta_i \leftarrow \nabla_{\theta_i}(r+\gamma[Q(s,(1-\epsilon) \pi_\phi(\cdot \mid s)+\epsilon \bar\pi_{\bar\phi}(\cdot \mid s))$\\
$\qquad -\alpha \mathcal{H}\cdot\pi_\phi(\cdot \mid s)]).$\\
\STATE $\phi \leftarrow \nabla_\phi[(1-\epsilon) \cdot \alpha \mathcal{H}\left(\pi_\phi(\cdot \mid \mathrm{s})\right)-Q(s, (1-\epsilon) $\\
$\qquad \pi_\phi(\cdot \mid \mathrm{s})+\epsilon \bar\pi_{\bar\phi}(\cdot \mid \mathrm{s})].$\\
\STATE $\bar\phi \leftarrow \nabla_{\bar\phi}\left[\epsilon \cdot Q(s, (1-\epsilon) \pi_\phi(\cdot \mid \mathrm{s})+\epsilon \bar\pi_{\bar\phi}(\cdot \mid \mathrm{s})\right].$\\
\STATE $\alpha \leftarrow \nabla_\alpha -\alpha\log\pi_{\phi}(\cdot\mid\,s)-\alpha\,\mathcal{H}_{0}.$\\
\STATE $\bar\theta_i \leftarrow \tau \theta_i+(1-\tau) \bar\theta_i.$\\
\ENDFOR
\ENDFOR
\STATE \textbf{Output}: $\theta_1,\theta_2,\phi ,\bar\phi $\\
\end{algorithmic}
\end{algorithm}
\subsection{Adaptive Perturbation for Adversarial Training : A case Study on SAC}
Now we apply our method to SAC algorithm. In A2P-SAC, the critic $Q_{\theta}(s_{t},a_{t})$ and the actor $\pi_{\phi}(s_t,a_t)$ in the original SAC algorithm are retained, and the adversarial policy $\bar{\pi}_{\bar{\phi}}(s_t,a_t)$ is added. The parameter $\bar{\theta}$ in the soft target $Q_{\bar{\theta}}(s_t,a_t)$ is obtained from the parameter $\theta$ in the exponentially moving average $Q_{\theta}$. Where the action here is obtained using the reparameterization trick, $a_t=f_{\phi}(\delta_{t};s_t)$, $\delta$ is noise. The same is true for the adversarial action $\bar{a}_t=\bar{f}_{\bar\phi}(\delta_{ t};s_t)$. We also use the auto-adjustable temperature parameter $\alpha$ from the original SAC algorithm. The updating of these network parameters is described in detail next.

In the policy evaluation step, the objective of critic is the Bellman residual between the current $Q$ and the target $Q$\cite{sutton2018reinforcement}. Target $Q$ is defined as
\begin{equation}
    \hat{Q}(s_t,a_{t}^{adpt})\stackrel {\Delta} {\mathop {=}}\,r(s_t,a_{t}^{adpt})+\gamma\mathbf{E}_{s_{t+1}\sim\,p}[\hat{V}(s_{t+1})],
\end{equation}
\begin{equation}
    \begin{aligned}
        \hat{V}(s_{t})\stackrel {\Delta} {\mathop {=}}\,&\min\limits_{{\bar{a}_t\sim\bar{\pi}(\cdot\mid\,s_t)}}\mathbf{E}_{a_t\sim\pi(\cdot\mid\,s_t)}[Q(s_t,(1-\epsilon^{adpt})a_t+\epsilon^{adpt}\bar{a}_t)\\&-(1-\epsilon^{adpt})\cdot\alpha\,\log\pi(a_t\mid\,s_t)].
    \end{aligned}
\end{equation}
Then, we train the critic $Q_{\theta}$ by minimizing the the Bellman residual
\begin{equation}
    \begin{aligned}
        J^{adpt}_{\epsilon-\mathcal{A}}(\theta)\stackrel {\Delta}{\mathop {=}}\,&\mathbf{E}_{s_t\sim\,\mathcal{D},\delta\sim\,\mathcal{N}}[\frac{1}{2}(Q_{\theta}(s_t,(1-\epsilon^{adpt})f_{\phi}(\delta_{t};s_t)\\&+\epsilon^{adpt}\bar{f}_{\bar\phi}(\delta_{t};s_t))-\hat{Q}_{\bar{\theta}}(s_t,(1-\epsilon^{adpt})f_{\phi}(\delta_{t};s_t)\\&+\epsilon^{adpt}\bar{f}_{\bar\phi}(\delta_{t};s_t)))^2].
    \end{aligned}
\end{equation}
In the policy improvement step, we train the actor by minimizing
\begin{equation}
    \begin{aligned}
        J^{adpt}_{\epsilon-A}(\phi)\stackrel{\Delta}{\mathop{=}}\,&\mathbf{E}_{s_t\sim\,\mathcal{D},\delta\sim\,\mathcal{N}}[(1-\epsilon^{adpt})\alpha\log\pi_{\phi}(f_{\phi}(\delta_{t};s_t)\mid\,s_t)\\&-Q_{\theta}(s_t,(1-\epsilon^{adpt})f_{\phi}(\delta_{t};s_t)\\&+\epsilon^{adpt}\bar{f}_{\bar\phi}(\delta_{t};s_t))].
    \end{aligned}
\end{equation}
Then we train the adversary by minimizing
\begin{equation}
    \begin{aligned}
        J^{adpt}_{\epsilon-A}(\bar\phi)\stackrel{\Delta}{\mathop{=}}\,&\mathbf{E}_{s_t\sim\,\mathcal{D},\delta\sim\,\mathcal{N}}[Q_{\theta}(s_t,(1-\epsilon^{adpt})f_{\phi}(\delta_{t};s_t)\\&+\epsilon^{adpt}\bar{f}_{\bar\phi}(\delta_{t};s_t))].
    \end{aligned}
\end{equation}
In the temperature $\alpha$ update step, we tune $\alpha$ by minimizing
\begin{equation}
    \begin{aligned}
        J_{\epsilon-A}^{adpt}(\alpha)\stackrel{\Delta}{\mathop{=}}\,&\mathbf{E}_{s_t\sim\,\mathcal{D},\delta\sim\,\mathcal{N}}[-\alpha\log\pi_{\phi}(f_{\phi}(\delta_{t};s_t)\mid\,s_t)\\&-\alpha\,\mathcal{H}_{0}],
    \end{aligned}
\end{equation}
where $\mathcal{H}$ is the target value of the entropy term. The detailed algorithm is shown in Algorithm~\ref{alg:algorithm}.

\begin{figure*}[htb]
    \centering
\subfigure{
\includegraphics[width=0.24\linewidth]{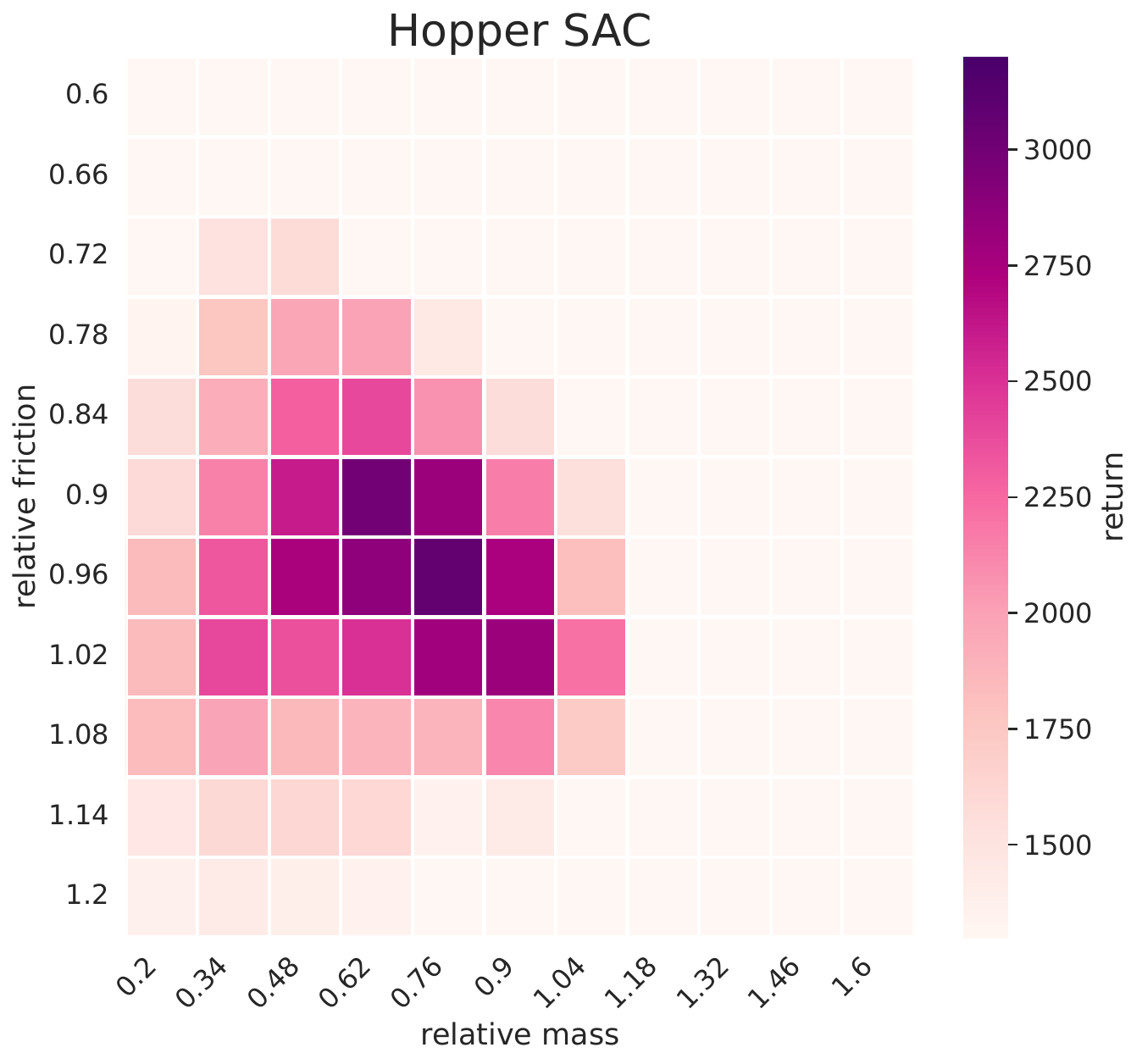}}
\hspace{-2mm}
\subfigure{\label{}
\includegraphics[width=0.24\linewidth]{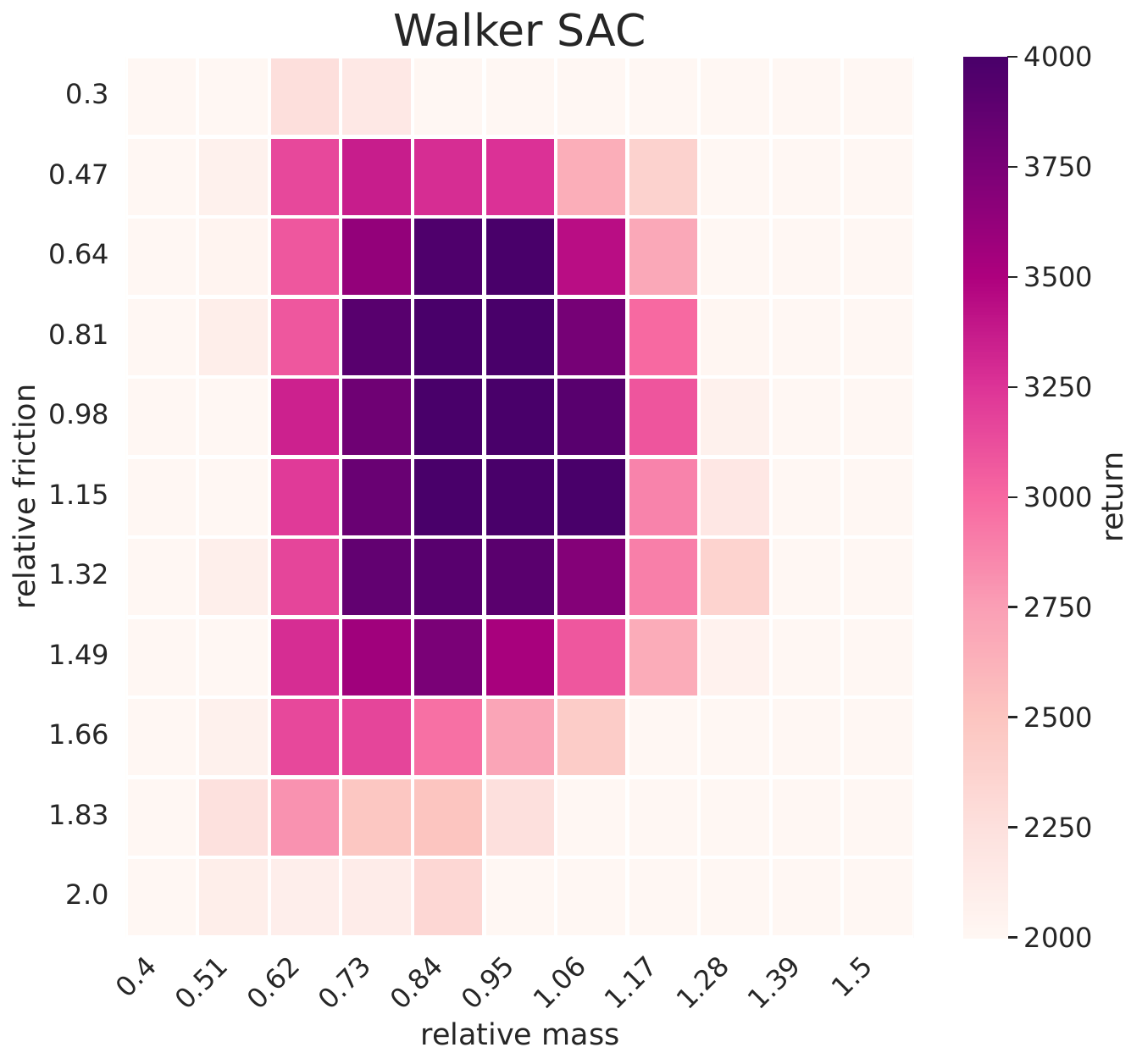}}
\hspace{-2mm}
\subfigure{
\includegraphics[width=0.24\linewidth]{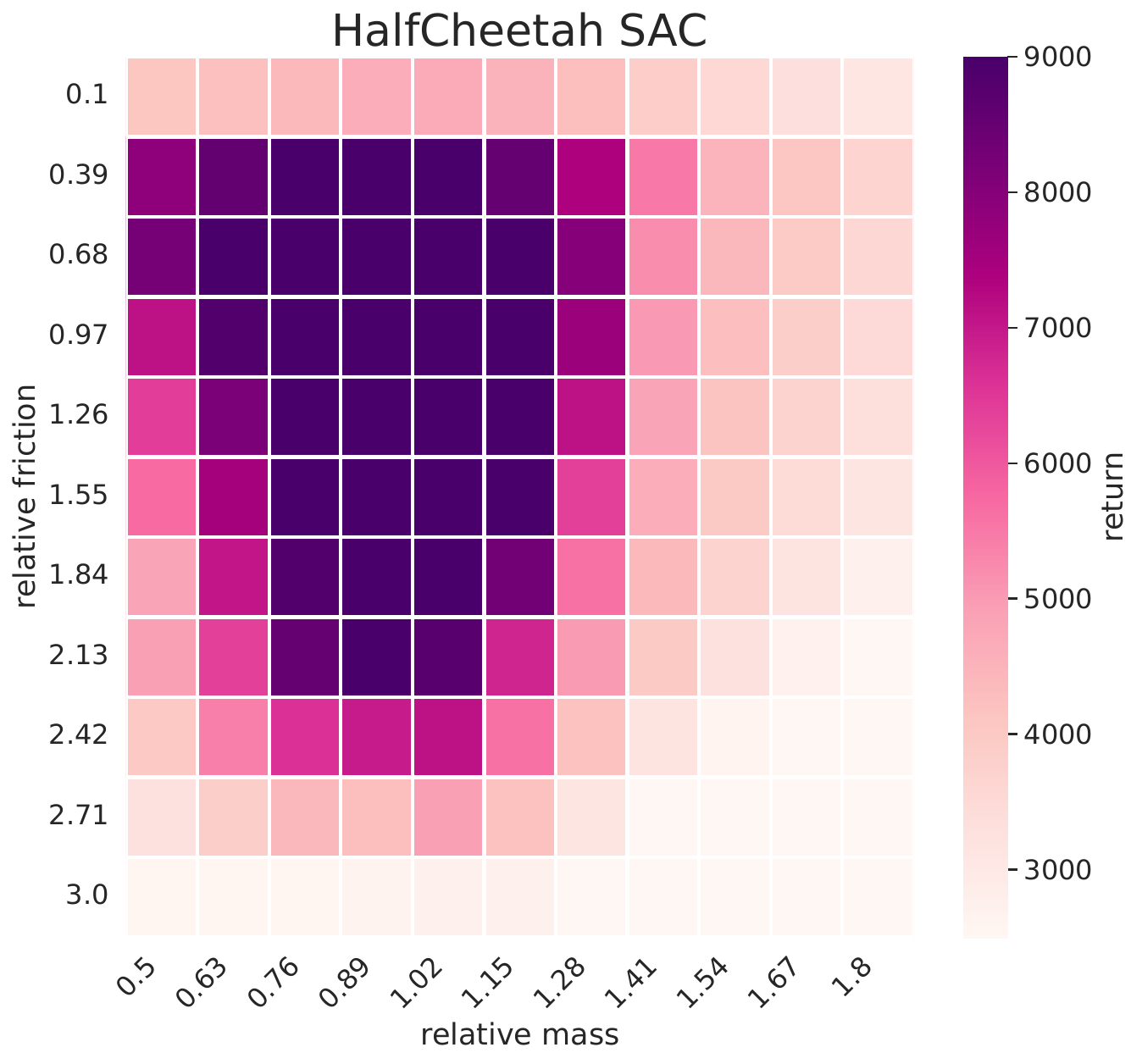}}
\hspace{-2mm}
\subfigure{\label{}
\includegraphics[width=0.24\linewidth]{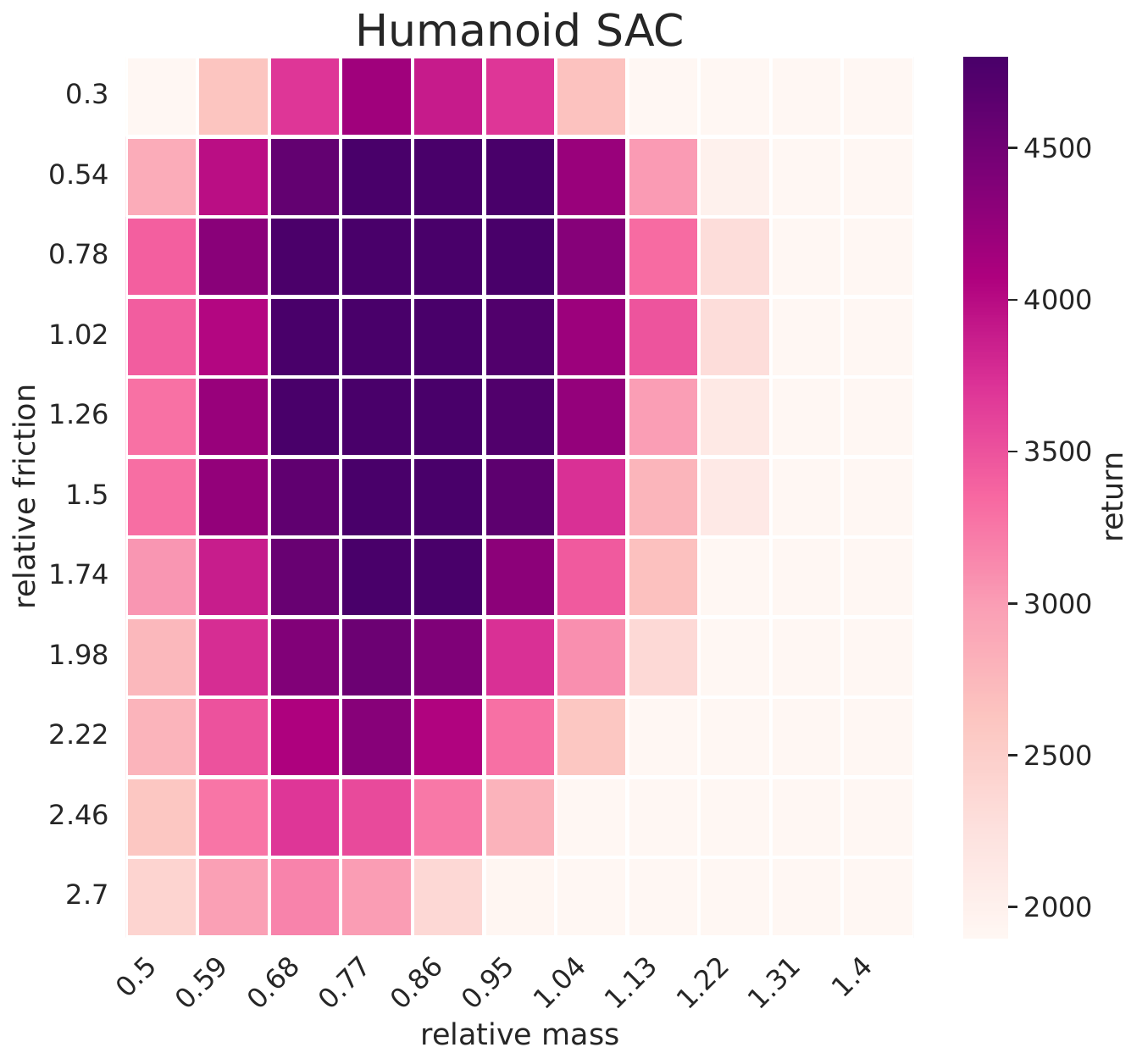}}
\subfigure{\label{}
\includegraphics[width=0.24\linewidth]{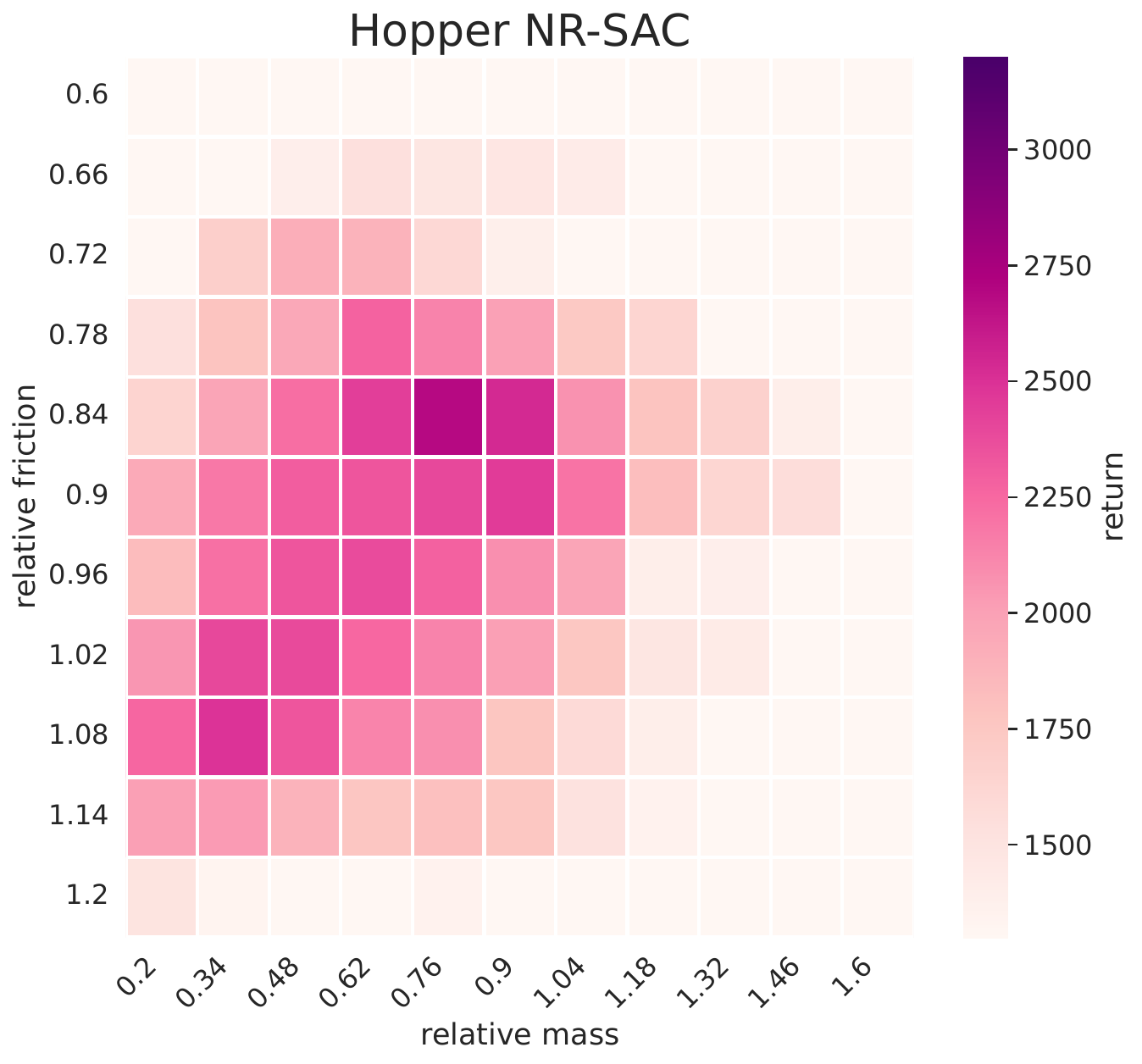}}
\hspace{-2mm}
\subfigure{\label{}
\includegraphics[width=0.24\linewidth]{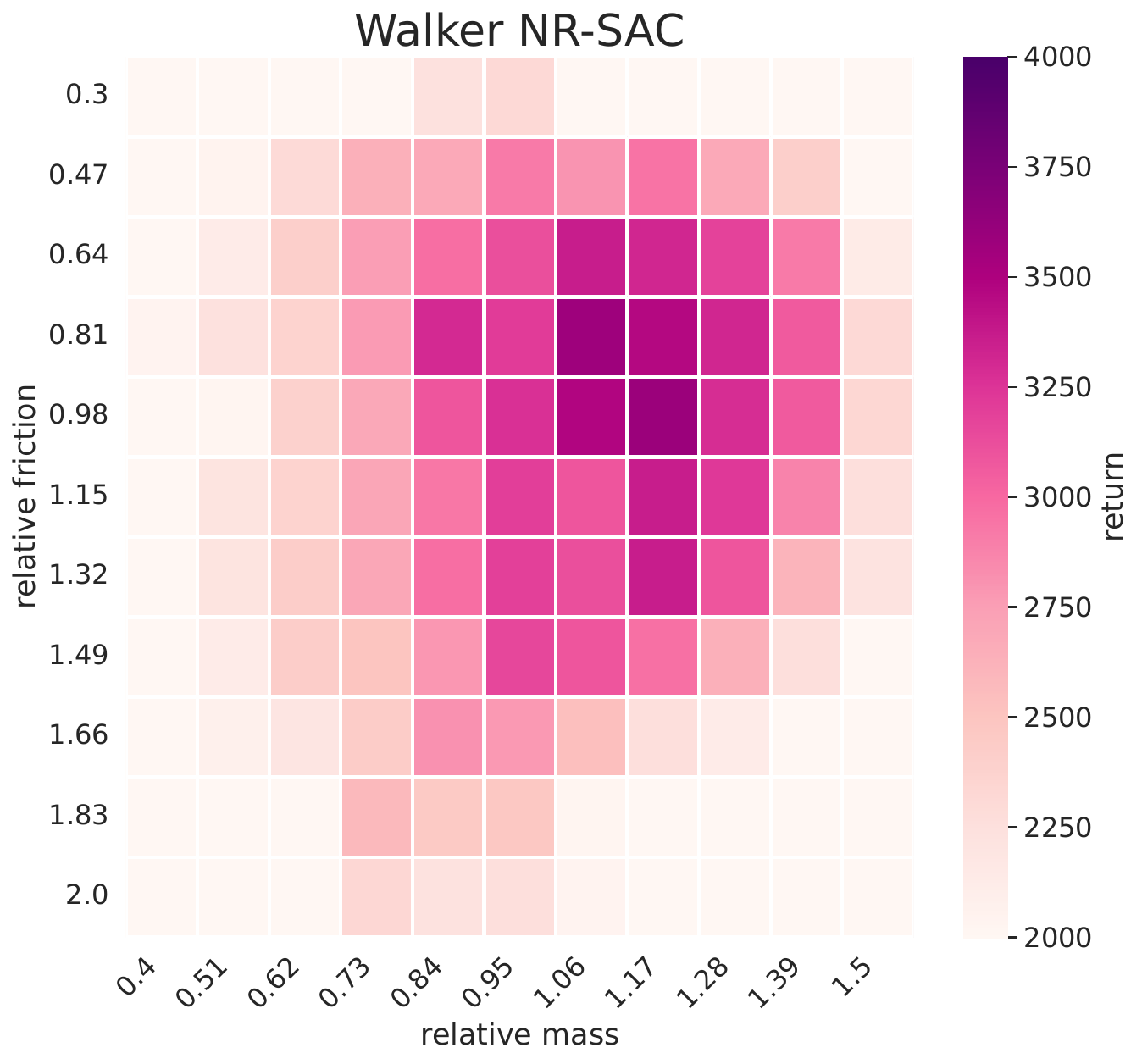}}
\hspace{-2mm}
\subfigure{
\includegraphics[width=0.24\linewidth]{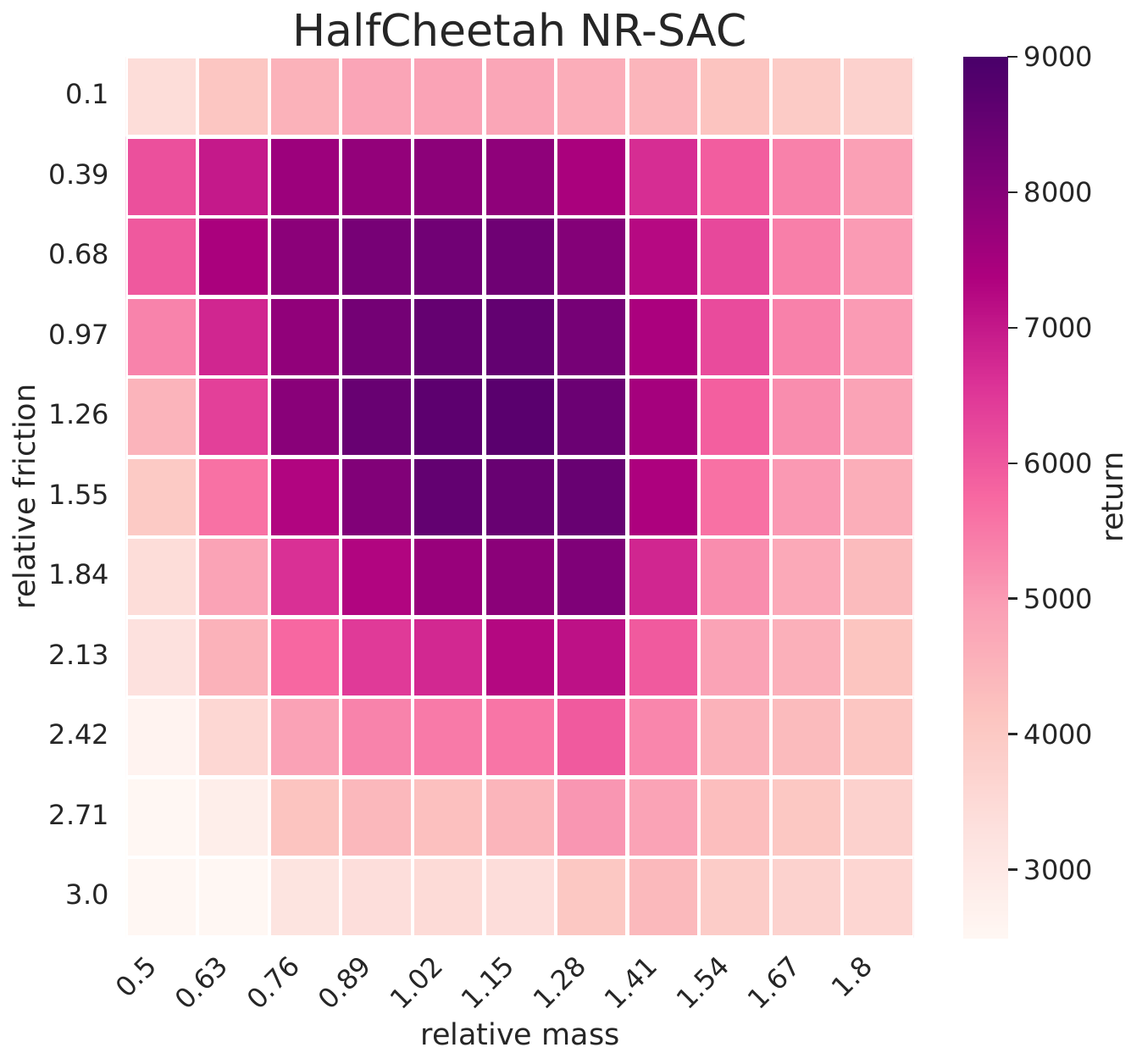}}
\hspace{-2mm}
\subfigure{\label{}
\includegraphics[width=0.24\linewidth]{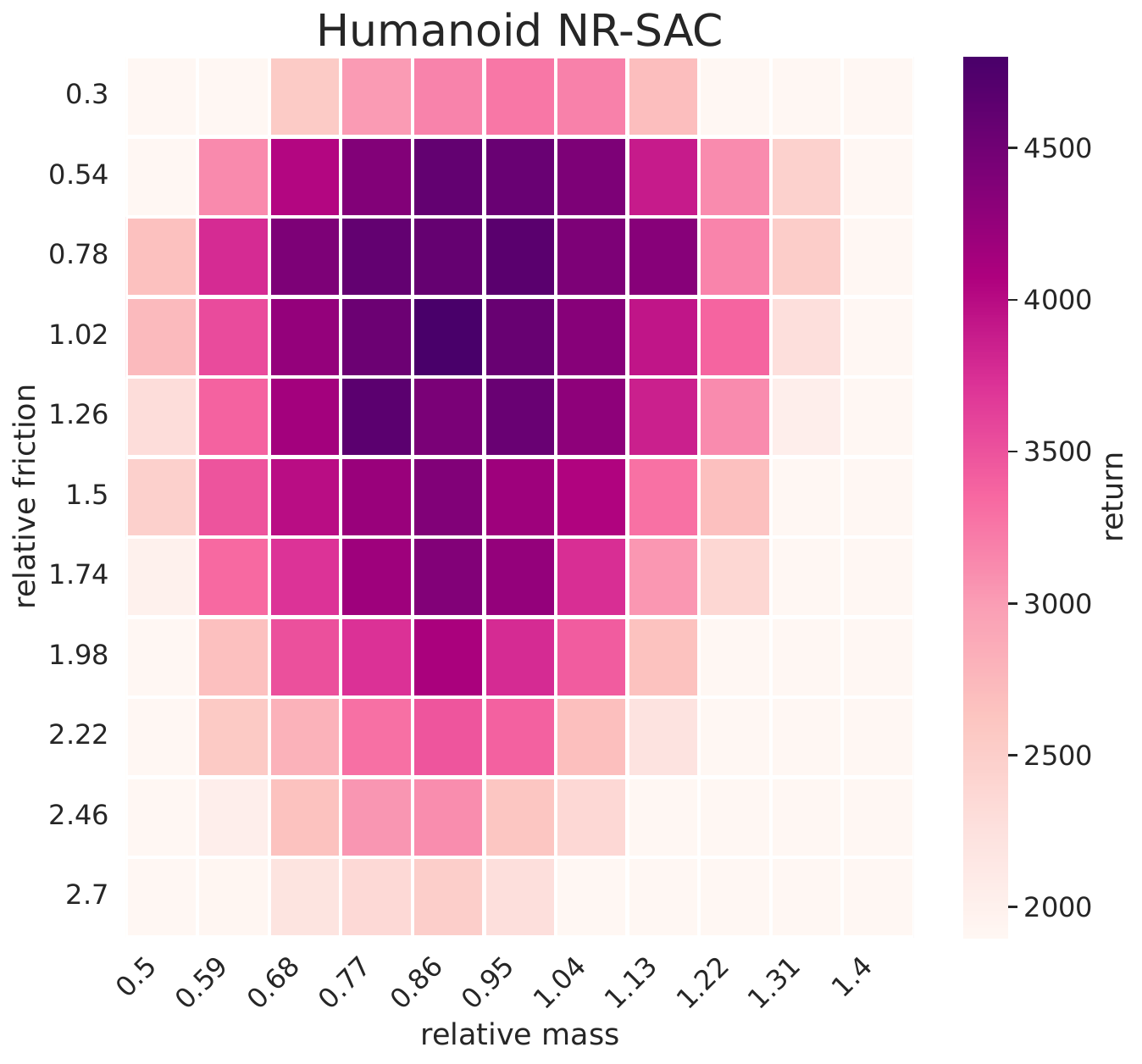}}
\subfigure{\label{}
\includegraphics[width=0.24\linewidth]{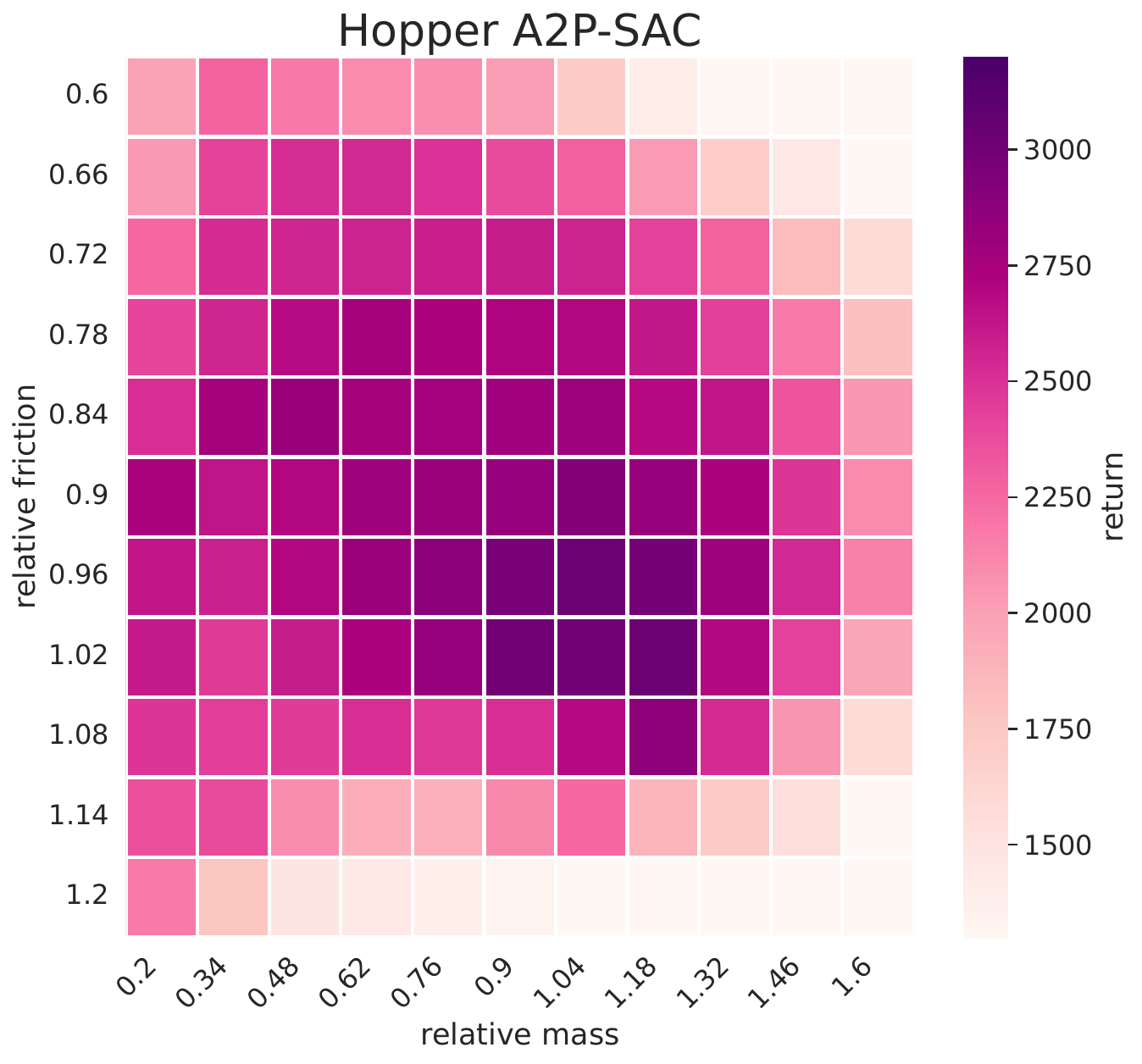}}
\hspace{-2mm}
\subfigure{\label{}
\includegraphics[width=0.24\linewidth]{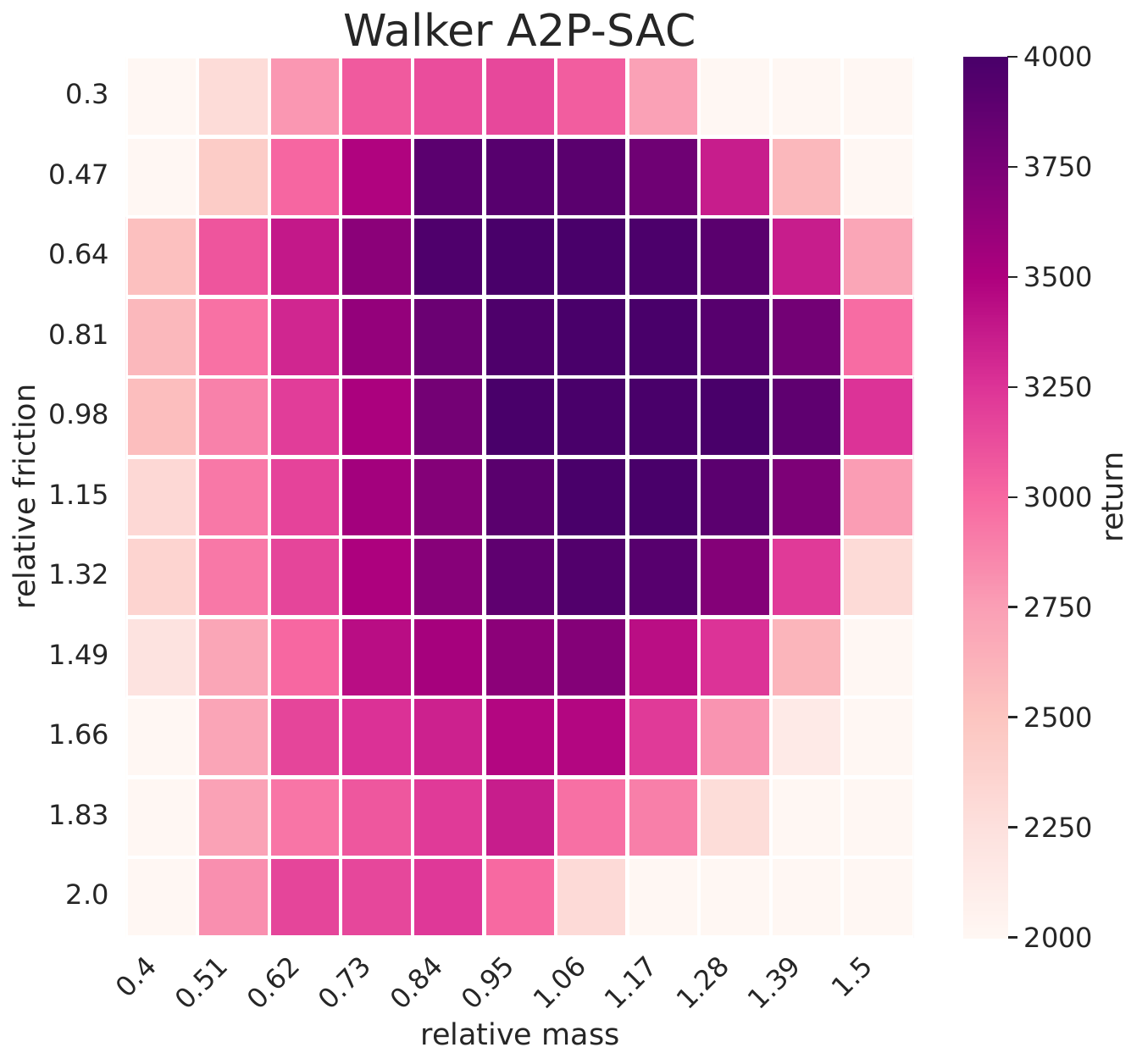}}
\hspace{-2mm}
\subfigure{
\includegraphics[width=0.24\linewidth]{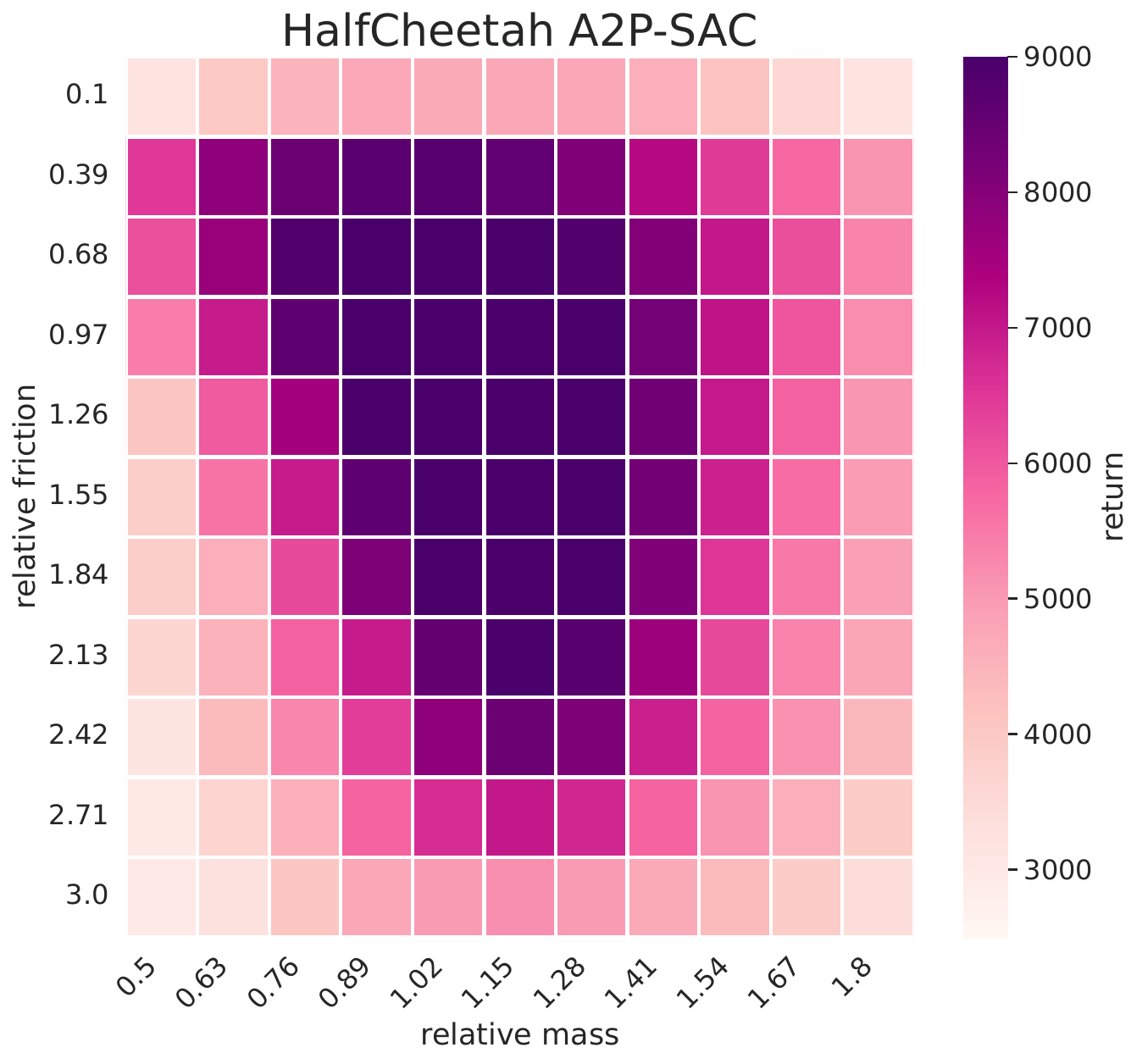}}
\hspace{-2mm}
\subfigure{\label{}
\includegraphics[width=0.24\linewidth]{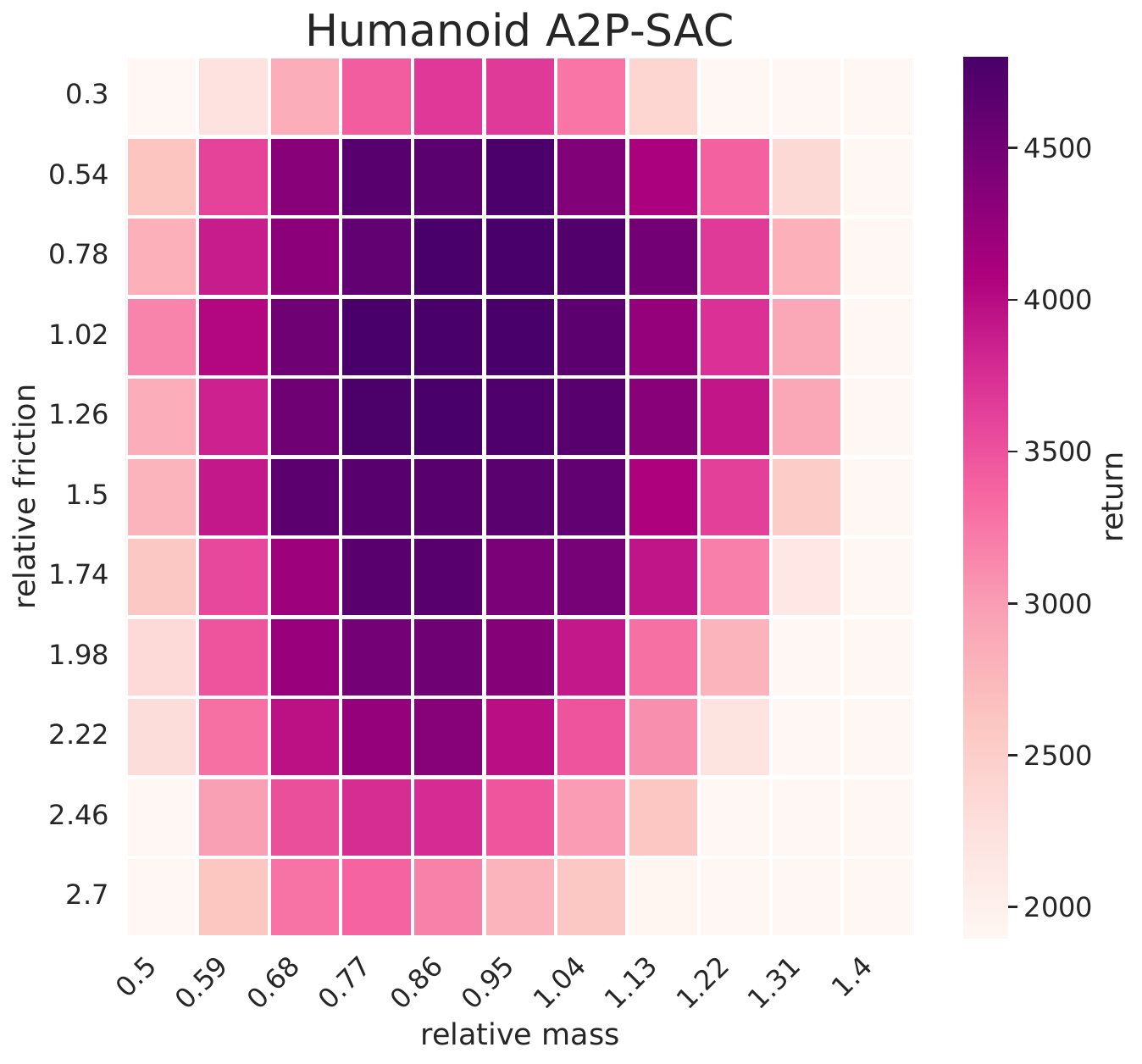}}
\vspace{-2mm}
\caption{Comparison of SAC, NR-SAC, and A2P-SAC in the target environment with perturbed mass and friction. For each task we randomly choose five seeds for training and save the last 10 episodes of each random seed. For each seed we randomly choose four policies for testing, then calculate the average performance. The results show that A2P-SAC can learn a more robust policy.}
\label{fig:mainrobust}
\end{figure*}
\section{Experiments}
In this section, we conduct experiments on the adaptive adversarial perturbations algorithm with SAC\cite{haarnoja2018soft}, which is robust and more widely used, in several MuJoCo benchmarks\cite{todorov2012mujoco} to evaluate its performance. It is more representative to use SAC as the baseline. In our paper, we conducted validation experiments based on NR-MDP\cite{tessler2019action}. Theoretically, we can also test it on other methods to combat perturbations.
\subsection{Implementation Details}
The hyperparameter $\epsilon$ serves as an adversarial perturbation coefficient. We search in the hopper-v3 task and set the initial value of $\epsilon$ to 0.1. Then we apply $\epsilon$ = 0.1 to all tasks. We normalize the results of SAC and A2P-SAC in all tasks. We keep all parameters in A2P-SAC the same as in the original SAC. We train 1000k steps (i.e., 1000 stages) in all tasks. We train the policy for each task with 5 random seeds. We perturb parameters (e.g., mass and friction) in the target environment to create perturbations in the transition dynamics.

To evaluate the effectiveness of A2P-SAC, we compared it with two baseline algorithms, namely vanilla SAC and NR-SAC. SAC is an off-policy reinforcement learning algorithm that combines policy gradient and maximum entropy principle, which can learn stable and efficient policies in continuous action spaces. NR-SAC is an algorithm that uses a fixed perturbation coefficient to perform adversarial training, which also has some achievements in improving robustness. We conducted experiments on four different tasks in the MuJoCo simulation environment, namely Hopper-v3, Walker-v3, HalfCheetah-v2 and Humanoid-v2. These environments have continuous action space and complex dynamics characteristics, which can effectively test the performance and robustness of the algorithms.
\begin{figure}[h]
    \centering 
    \vspace{-4mm}
    \begin{minipage}{0.49\linewidth}
        \centering
 \includegraphics[width=0.9\linewidth]{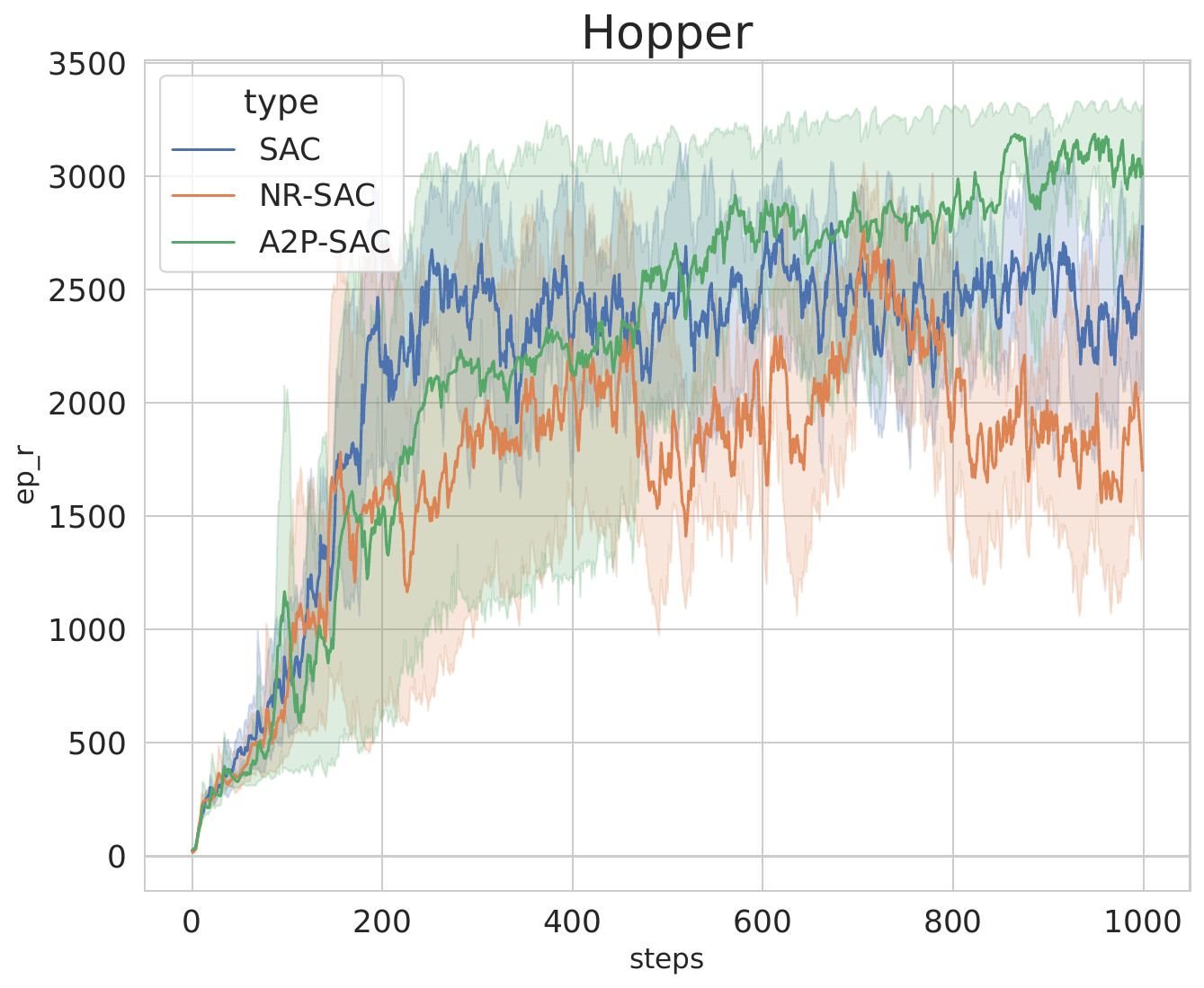}
    \end{minipage}
    \hspace{-5mm}
 \begin{minipage}{0.49\linewidth}
        \centering
 \includegraphics[width=0.9\linewidth]{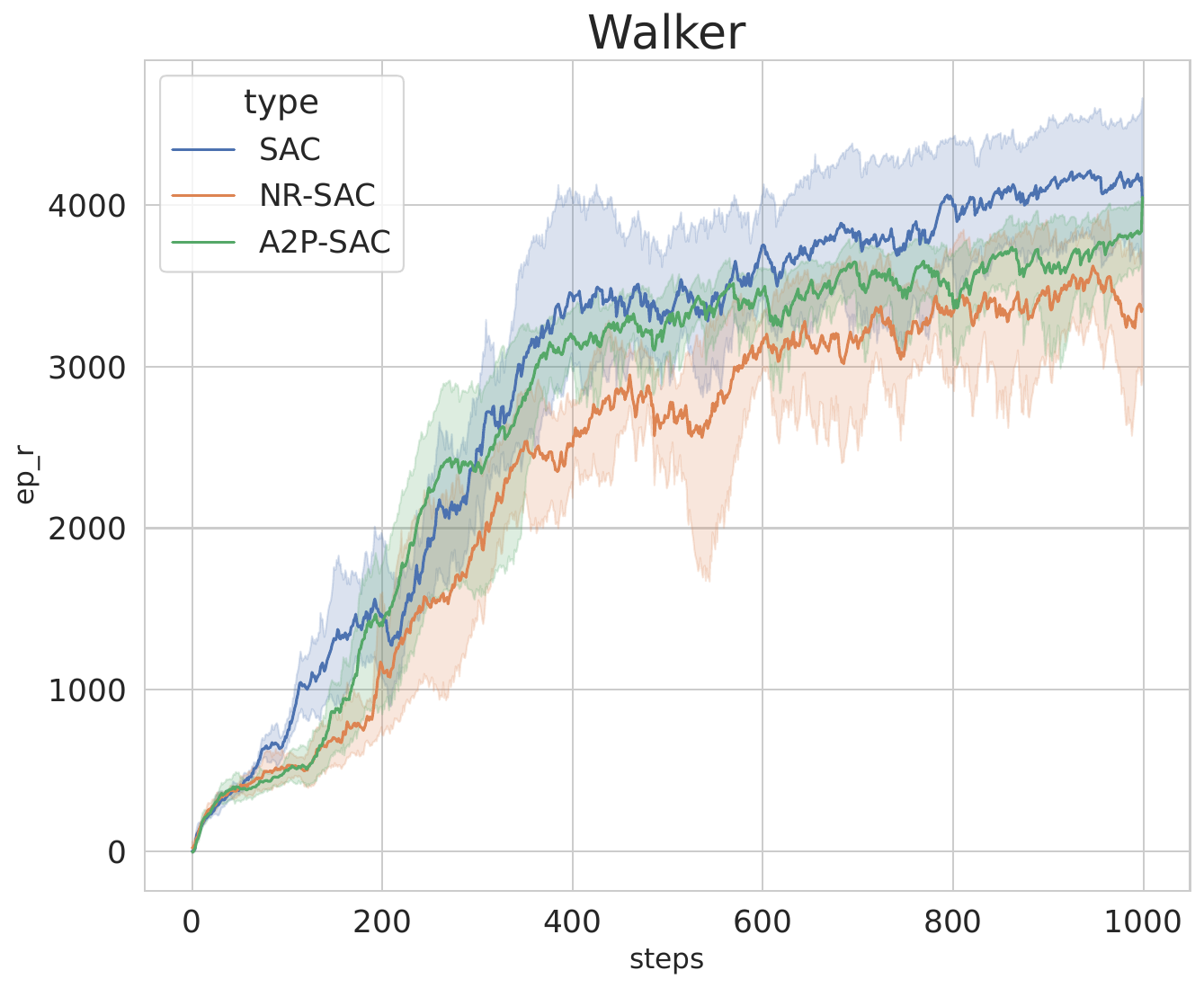}
    \end{minipage}
\begin{minipage}{0.49\linewidth}
        \centering
 \includegraphics[width=0.9\linewidth]{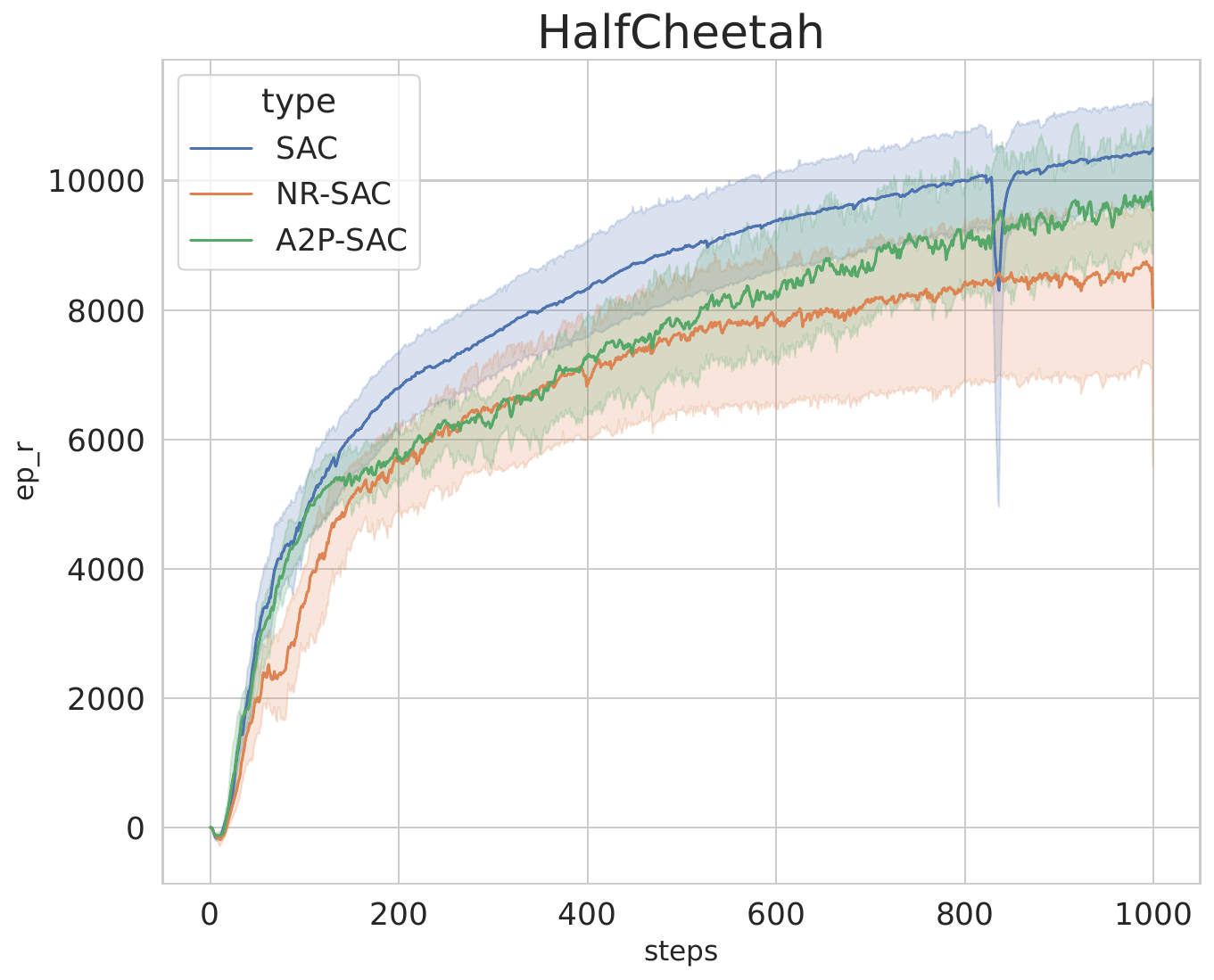}
    \end{minipage}
    \hspace{-5mm}
 \begin{minipage}{0.49\linewidth}
        \centering
\includegraphics[width=0.9\linewidth]{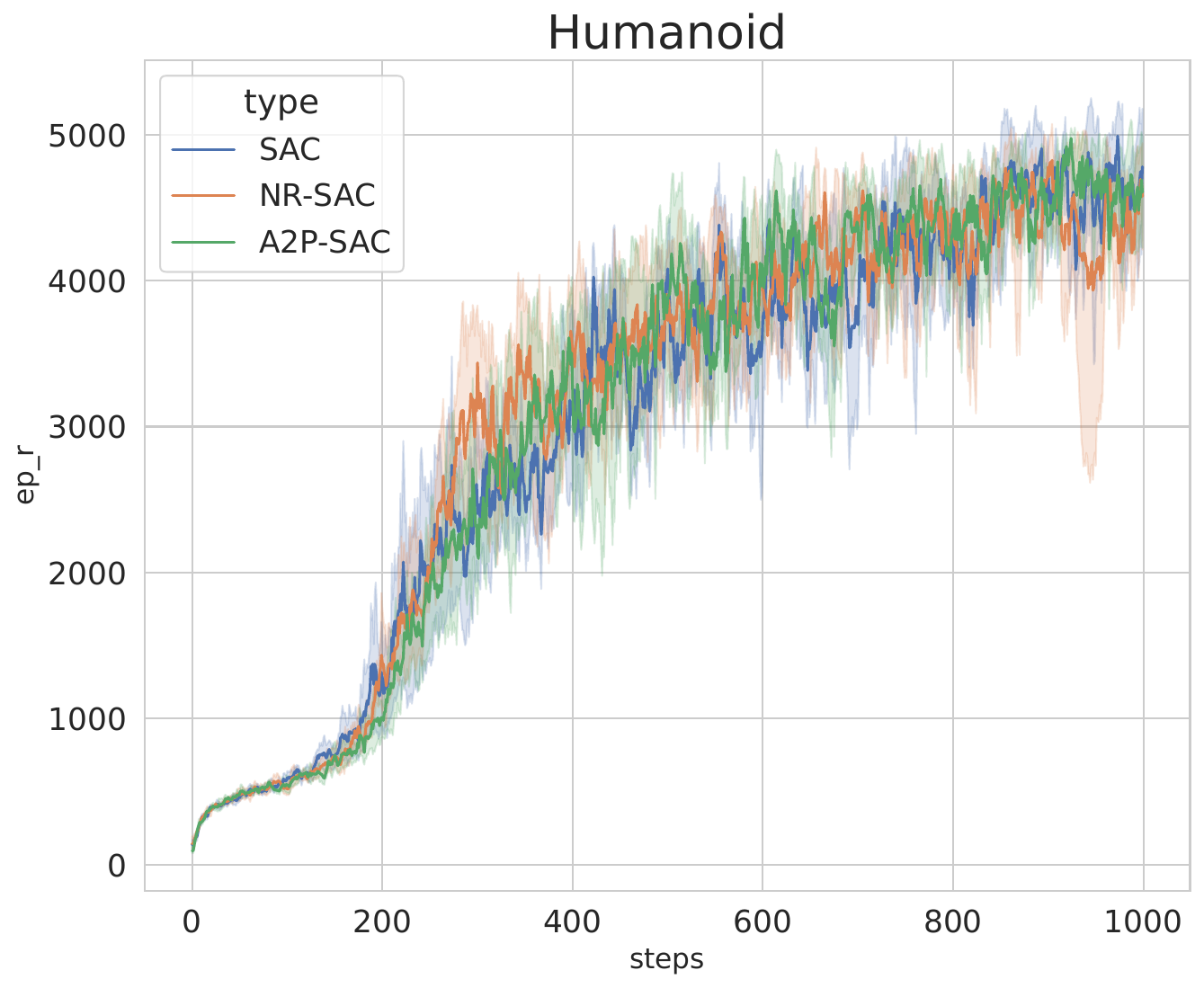}
    \end{minipage}
    \vspace{-2mm}
\caption{Comparison of training curves for different algorithms. For each algorithm in each task we train 1000k steps. The results show that our algorithm can approximate the performance of SAC in most tasks and is more stable than NR-SAC.}
\label{fig:train}
\vspace{-4mm}
\end{figure}
\subsection{Performance Evaluation}
\begin{figure}[h]
    \centering
    \vspace{-3mm}
\subfigure{
\includegraphics[width=0.45\linewidth]{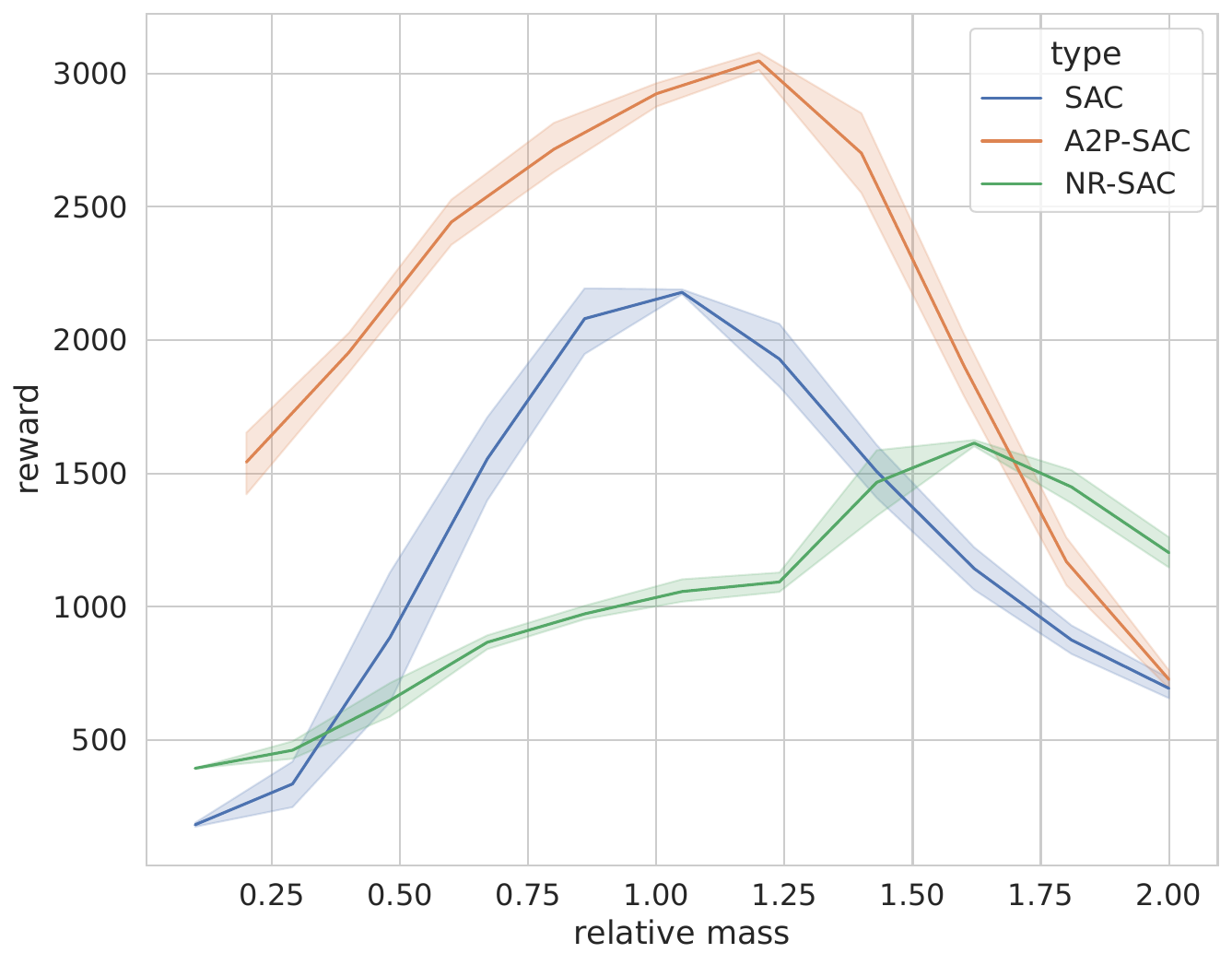}}
\hspace{-2mm}
\subfigure{\label{}
\includegraphics[width=0.45\linewidth]{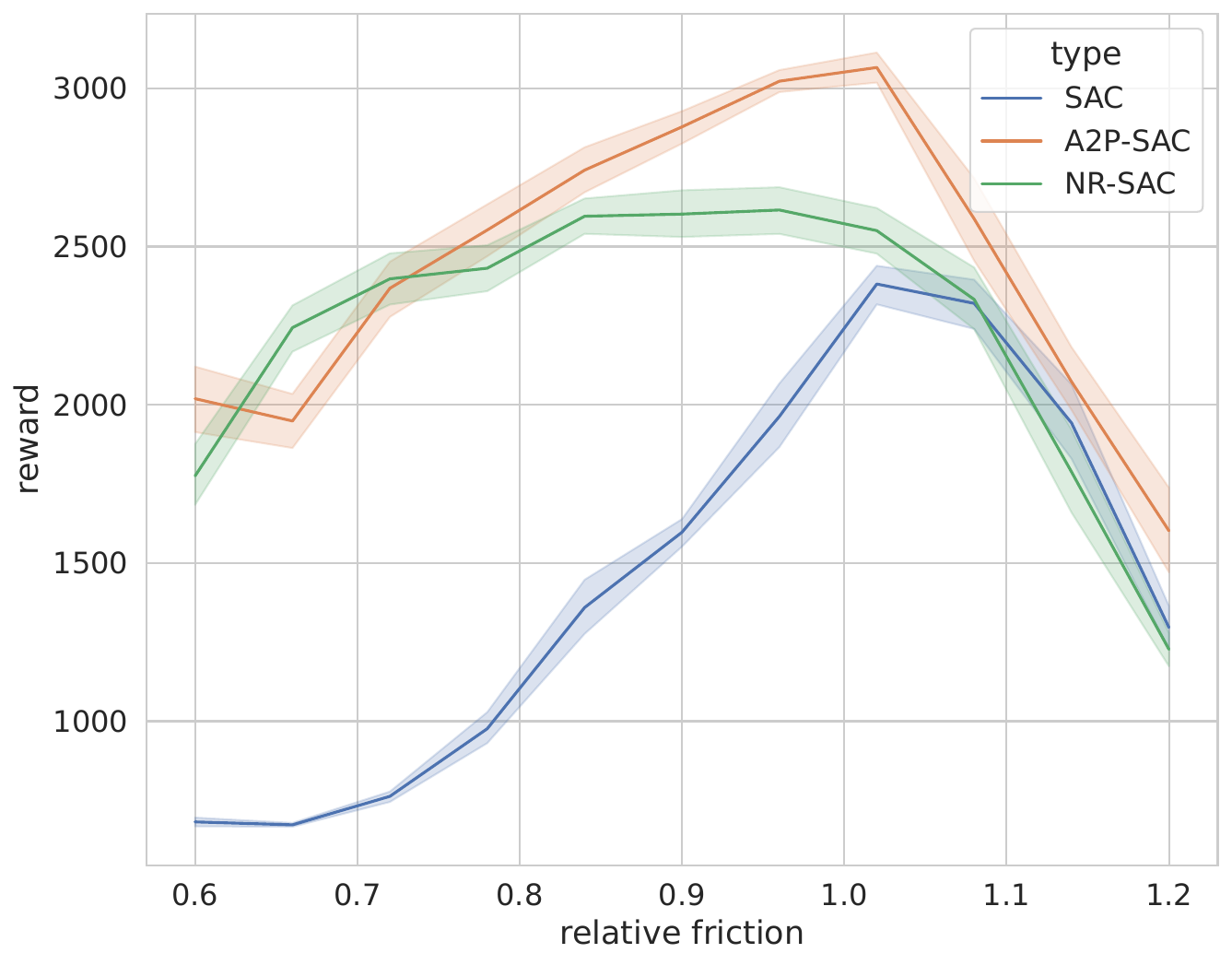}}
\vspace{-2mm}
\caption{Comparing the robustness of SAC, NR-SAC and A2P-SAC with one perturbed parameter(mass or friction) in target environment. The higher the reward, the better the performance. If the model performs well in multiple test environments, it means that the model has strong robustness. The results show that A2P method learns a more robust policy than others.}
\label{fig:changeone}
\vspace{-3mm}
\end{figure}

\textbf{Comparative Evaluation}
Figure~\ref{fig:train} shows the training curves of the three algorithms on the four tasks. From the figure, we can see that A2P-SAC can well approximate the performance of the original SAC, while NR-SAC shows lower training performance. This is consistent with our expectation, because the adversarial policy will interfere with the learning of the robust policy, resulting in a decrease in the cumulative returns. Moreover, using the adaptive perturbation method can alleviate the dilemma of return reduction to some extent compared to using the fixed perturbation method. Secondly, this perturbation can also improve the generalization ability of the robust policy, enabling it to adapt to different environmental parameters. To verify this, we show the test results of the three algorithms under varying environmental parameters in Figure~\ref{fig:changeone}, where the x-axis represents the range of environmental parameter changes and the y-axis represents the average reward. We changed the mass or friction coefficient of the environment to simulate different perturbation scenarios. From the figure, we can see that our method A2P-SAC outperforms the other two methods in most environments, indicating that our method has some effectiveness.

\textbf{Robustness against action space}
Real tasks may be complex, and they cannot be simply simulated by changing one parameter. Therefore, we simultaneously change multiple parameters (mass and friction) in the environment to simulate the perturbation of the real environment. In Figure~\ref{fig:mainrobust}, we test the robustness of the three algorithms on four different tasks. Each grid represents an environment. And in our experiment, we simulate 121 different perturbed environments for each task, and the return value of each grid is obtained by averaging 160 calculations. The darker the color in the grid, the higher the performance, and the more dark-colored grids, the greater the robustness. From the figure, we can see that our method A2P outperforms the other two methods in all four tasks, especially in Hopper-v3 and Walker-v3. This shows that our method A2P can indeed learn a more robust policy, and our method has strong robustness and adaptability. This also demonstrates that the A2P method can effectively use adaptive perturbation to improve the quality of the robust policy, rather than simply ignoring or suppressing the impact of the adversarial policy.

\subsection{Ablation Study}
\begin{figure}
    \centering
    \vspace{-5mm}
\subfigure{
\includegraphics[width=0.45\linewidth]{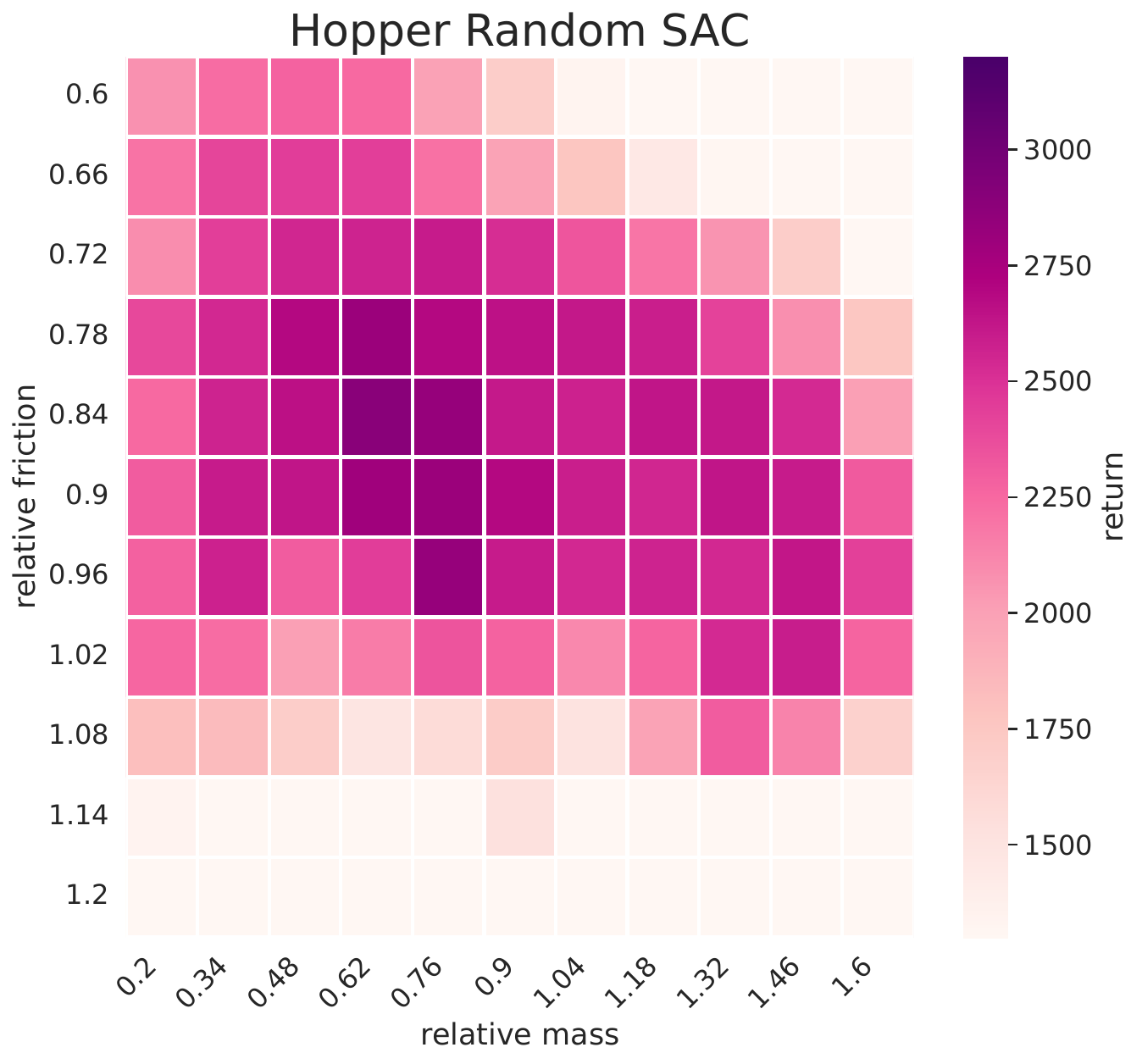}}
\vspace{-4mm}
\hspace{-2mm}
\subfigure{\label{}
\includegraphics[width=0.45\linewidth]{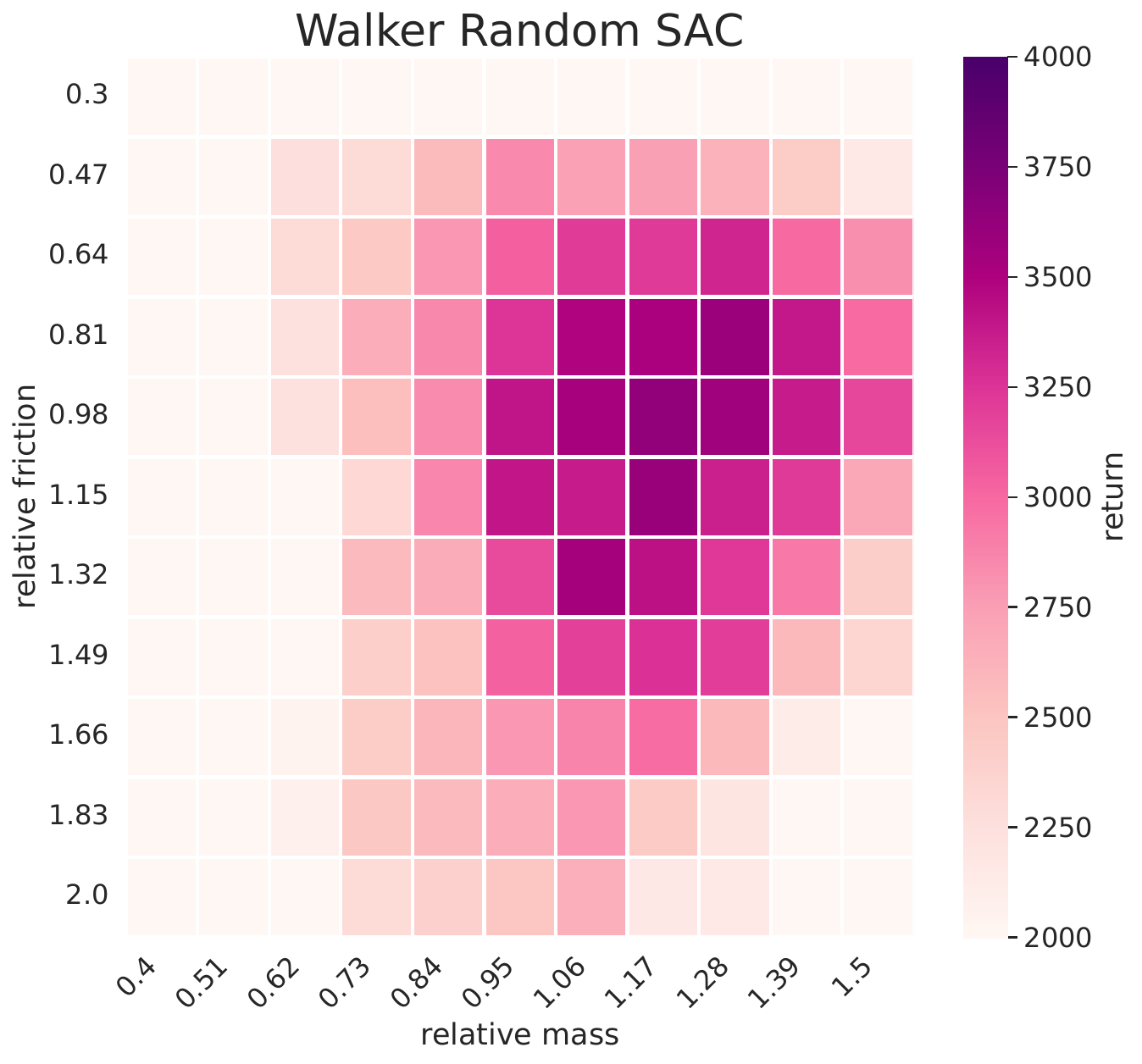}}
\vspace{-1mm}
\subfigure{\label{}
\includegraphics[width=0.45\linewidth]{figure/Hopper_adaptive_mdp_impro_5.pdf}}
\vspace{-1mm}
\hspace{-2mm}
\subfigure{\label{}
\includegraphics[width=0.45\linewidth]{figure/Walker_adaptive_mdp_impro_5.pdf}}
\vspace{-2mm}
\caption{Ablation study for A2P method with random and adaptive perturbations. We keep the setting same in two methods. The darker the color in the grid, the higher the performance, and the more dark-colored grids, the greater the robustness. The results show that A2P methods are more capable of learning a robust policy than randomized perturbations.}
\label{fig:ablation}
\vspace{-4mm}
\end{figure}

\textbf{The importance of adaptive perturbations}
Here we discuss and present a method of performing an ablation study between random perturbations and adaptive perturbations on the current performance. Random perturbation means choosing a coefficient $\epsilon$ randomly within a suitable range in each training epoch, to make small changes to the model parameters. Adaptive perturbation means calculating a coefficient $\epsilon$ that matches the model performance by measuring the factors in each training epoch, to make targeted changes to the model parameters. Figure~\ref{fig:ablation} shows that our method is more robust than the random method in the test environment. This suggests that adaptive perturbation acquired effective information, which can achieve a balance between the model's optimal performance in the source environment and its performance in the target environment, and mitigate the robustness challenge of reinforcement learning to a certain degree.
\begin{table}[ht]
    \centering
    \vspace{-4mm}
    \caption{Average episode rewards $\pm$ standard deviation on A2P-SAC in Hopper-v3 task.}
    \begin{tabular}{ ccc } 
 \toprule
  Parameter $\beta$ & Natural Reward & Attack Reward(state) \\ 
 \midrule
 0 & $2677.9\pm856.7$ & $1589.9\pm238.6$ \\ 
 \midrule
 0.1 & $2857.7\pm350.1$ & $2420.4\pm725.5$ \\ 
 \midrule
 0.3 & $2736.8\pm300.0$ & $2291.9\pm776.8$ \\ 
 \midrule
 \rowcolor{black!30}
  0.5 & $3042.8\pm476.7$ & $2620.7\pm805.5$ \\ 
 \midrule
  0.7 & $2893.8\pm308.3$ & $2328.3\pm826.2$ \\ 
 \midrule
  1.0 & $2972.2\pm536.8$ & $2565.6\pm791.1$ \\ 
 \bottomrule
\end{tabular}
    \label{tab:my_label1}
\vspace{-4mm}
\end{table}

\textbf{Sensitivity Analysis on $\beta$}
To study the impact of the parameter $\beta$ in our method on the model performance, we conducted a parameter sensitivity analysis experiment. The parameter $\beta$ is used to balance the weights of the historical and current action distances in the training process, which affects the size and direction of the adaptive perturbation. We show the reward of the A2P method under different $\beta$ values in Table~\ref{tab:my_label1}, under natural reward (no attack) and adversarial state attack (observation perturbation) scenarios. We choose $\beta$ as 0.1, 0.3, 0.5, 0.7, 1.0 for the experiment. The results show that the reward of the A2P method does not change much under different values, indicating that the A2P method has strong robustness to $\beta$. At the same time, we find that the gray row corresponding to $\beta$ value of 0.5 has the highest reward of the A2P method, while indicating that this is a reasonable weight balance point. Therefore, we set it as 0.5 as the optimal parameter of the algorithm.

\section{Conclusion}
In this paper, we present a comprehensive analysis of the impact of perturbations on the effectiveness of adversarial training in the context of adversarial reinforcement learning. We also discuss the trade-off between achieving robustness in the target environment and maintaining high average performance in the source environment. Furthermore, we demonstrate the advantages of our approach in quantifying and handling the dynamic uncertainty of policy transitions. Motivated by these analyses, we propose an adaptive adversarial perturbation method that operates in the action-space to mitigate this trade-off. We conduct extensive experiments on various MuJoCo tasks and show that our adaptive adversarial perturbation method can achieve stable performance and greater robustness than existing approaches.

\section{Acknowledgments}
The authors would like to thank all the anonymous reviewers for their insightful comments. This work was supported by National Key R\&D Program of China under contract 2022ZD0119801 and National Nature Science Foundations of China grants U23A20388, U19B2026, U19B2044, and 62021001.

\bibliographystyle{IEEEtran}
\footnotesize
\bibliography{IEEEexample}

\begin{thebibliography}{10}
\providecommand{\url}[1]{#1}
\csname url@samestyle\endcsname
\providecommand{\newblock}{\relax}
\providecommand{\bibinfo}[2]{#2}
\providecommand{\BIBentrySTDinterwordspacing}{\spaceskip=0pt\relax}
\providecommand{\BIBentryALTinterwordstretchfactor}{4}
\providecommand{\BIBentryALTinterwordspacing}{\spaceskip=\fontdimen2\font plus
\BIBentryALTinterwordstretchfactor\fontdimen3\font minus
  \fontdimen4\font\relax}
\providecommand{\BIBforeignlanguage}[2]{{%
\expandafter\ifx\csname l@#1\endcsname\relax
\typeout{** WARNING: IEEEtran.bst: No hyphenation pattern has been}%
\typeout{** loaded for the language `#1'. Using the pattern for}%
\typeout{** the default language instead.}%
\else
\language=\csname l@#1\endcsname
\fi
#2}}
\providecommand{\BIBdecl}{\relax}
\BIBdecl

\bibitem{mnih2015human}
V.~Mnih, K.~Kavukcuoglu, D.~Silver, A.~A. Rusu, J.~Veness, M.~G. Bellemare,
  A.~Graves, M.~Riedmiller, A.~K. Fidjeland, G.~Ostrovski \emph{et~al.},
  ``Human-level control through deep reinforcement learning,'' \emph{nature},
  vol. 518, no. 7540, pp. 529--533, 2015.

\bibitem{lillicrap2015continuous}
T.~P. Lillicrap, J.~J. Hunt, A.~Pritzel, N.~Heess, T.~Erez, Y.~Tassa,
  D.~Silver, and D.~Wierstra, ``Continuous control with deep reinforcement
  learning,'' \emph{arXiv preprint arXiv:1509.02971}, 2015.

\bibitem{vinyals2019grandmaster}
O.~Vinyals, I.~Babuschkin, W.~M. Czarnecki, M.~Mathieu, A.~Dudzik, J.~Chung,
  D.~H. Choi, R.~Powell, T.~Ewalds, P.~Georgiev \emph{et~al.}, ``Grandmaster
  level in starcraft ii using multi-agent reinforcement learning,''
  \emph{Nature}, vol. 575, no. 7782, pp. 350--354, 2019.

\bibitem{levine2016end}
S.~Levine, C.~Finn, T.~Darrell, and P.~Abbeel, ``End-to-end training of deep
  visuomotor policies,'' \emph{The Journal of Machine Learning Research},
  vol.~17, no.~1, pp. 1334--1373, 2016.

\bibitem{andrychowicz2020learning}
O.~M. Andrychowicz, B.~Baker, M.~Chociej, R.~Jozefowicz, B.~McGrew,
  J.~Pachocki, A.~Petron, M.~Plappert, G.~Powell, A.~Ray \emph{et~al.},
  ``Learning dexterous in-hand manipulation,'' \emph{The International Journal
  of Robotics Research}, vol.~39, no.~1, pp. 3--20, 2020.

\bibitem{sangiovanni2018deep}
B.~Sangiovanni, A.~Rendiniello, G.~P. Incremona, A.~Ferrara, and M.~Piastra,
  ``Deep reinforcement learning for collision avoidance of robotic
  manipulators,'' in \emph{2018 European Control Conference (ECC)}.\hskip 1em
  plus 0.5em minus 0.4em\relax IEEE, 2018, pp. 2063--2068.

\bibitem{sallab2017deep}
A.~E. Sallab, M.~Abdou, E.~Perot, and S.~Yogamani, ``Deep reinforcement
  learning framework for autonomous driving,'' \emph{arXiv preprint
  arXiv:1704.02532}, 2017.

\bibitem{aradi2020survey}
S.~Aradi, ``Survey of deep reinforcement learning for motion planning of
  autonomous vehicles,'' \emph{IEEE Transactions on Intelligent Transportation
  Systems}, vol.~23, no.~2, pp. 740--759, 2020.

\bibitem{jiang2021monotonic}
Y.~Jiang, C.~Li, W.~Dai, J.~Zou, and H.~Xiong, ``Monotonic robust policy
  optimization with model discrepancy,'' in \emph{International Conference on
  Machine Learning}.\hskip 1em plus 0.5em minus 0.4em\relax PMLR, 2021, pp.
  4951--4960.

\bibitem{tzeng2015towards}
E.~Tzeng, C.~Devin, J.~Hoffman, C.~Finn, X.~Peng, S.~Levine, K.~Saenko, and
  T.~Darrell, ``Towards adapting deep visuomotor representations from simulated
  to real environments,'' \emph{arXiv preprint arXiv:1511.07111}, vol.~2,
  no.~3, 2015.

\bibitem{peng2018sim}
X.~B. Peng, M.~Andrychowicz, W.~Zaremba, and P.~Abbeel, ``Sim-to-real transfer
  of robotic control with dynamics randomization,'' in \emph{2018 IEEE
  international conference on robotics and automation (ICRA)}.\hskip 1em plus
  0.5em minus 0.4em\relax IEEE, 2018, pp. 3803--3810.

\bibitem{iyengar2005robust}
G.~N. Iyengar, ``Robust dynamic programming,'' \emph{Mathematics of Operations
  Research}, vol.~30, no.~2, pp. 257--280, 2005.

\bibitem{rajeswaran2016epopt}
A.~Rajeswaran, S.~Ghotra, B.~Ravindran, and S.~Levine, ``Epopt: Learning robust
  neural network policies using model ensembles,'' \emph{arXiv preprint
  arXiv:1610.01283}, 2016.

\bibitem{pinto2017robust}
L.~Pinto, J.~Davidson, R.~Sukthankar, and A.~Gupta, ``Robust adversarial
  reinforcement learning,'' in \emph{International Conference on Machine
  Learning}.\hskip 1em plus 0.5em minus 0.4em\relax PMLR, 2017, pp. 2817--2826.

\bibitem{zhang2020robust}
H.~Zhang, H.~Chen, C.~Xiao, B.~Li, M.~Liu, D.~Boning, and C.-J. Hsieh, ``Robust
  deep reinforcement learning against adversarial perturbations on state
  observations,'' \emph{Advances in Neural Information Processing Systems},
  vol.~33, pp. 21\,024--21\,037, 2020.

\bibitem{tessler2019action}
C.~Tessler, Y.~Efroni, and S.~Mannor, ``Action robust reinforcement learning
  and applications in continuous control,'' in \emph{International Conference
  on Machine Learning}.\hskip 1em plus 0.5em minus 0.4em\relax PMLR, 2019, pp.
  6215--6224.

\bibitem{tobin2017domain}
J.~Tobin, R.~Fong, A.~Ray, J.~Schneider, W.~Zaremba, and P.~Abbeel, ``Domain
  randomization for transferring deep neural networks from simulation to the
  real world,'' in \emph{2017 IEEE/RSJ international conference on intelligent
  robots and systems (IROS)}.\hskip 1em plus 0.5em minus 0.4em\relax IEEE,
  2017, pp. 23--30.

\bibitem{yang2017convex}
I.~Yang, ``A convex optimization approach to distributionally robust markov
  decision processes with wasserstein distance,'' \emph{IEEE control systems
  letters}, vol.~1, no.~1, pp. 164--169, 2017.

\bibitem{mankowitz2019robust}
D.~J. Mankowitz, N.~Levine, R.~Jeong, Y.~Shi, J.~Kay, A.~Abdolmaleki, J.~T.
  Springenberg, T.~Mann, T.~Hester, and M.~Riedmiller, ``Robust reinforcement
  learning for continuous control with model misspecification,'' \emph{arXiv
  preprint arXiv:1906.07516}, 2019.

\bibitem{abdullah2019wasserstein}
M.~A. Abdullah, H.~Ren, H.~B. Ammar, V.~Milenkovic, R.~Luo, M.~Zhang, and
  J.~Wang, ``Wasserstein robust reinforcement learning,'' \emph{arXiv preprint
  arXiv:1907.13196}, 2019.

\bibitem{kour2014real}
G.~Kour and R.~Saabne, ``Real-time segmentation of on-line handwritten arabic
  script,'' in \emph{Frontiers in Handwriting Recognition (ICFHR), 2014 14th
  International Conference on}.\hskip 1em plus 0.5em minus 0.4em\relax IEEE,
  2014, pp. 417--422.

\bibitem{madry2017towards}
A.~Madry, A.~Makelov, L.~Schmidt, D.~Tsipras, and A.~Vladu, ``Towards deep
  learning models resistant to adversarial attacks,'' \emph{arXiv preprint
  arXiv:1706.06083}, 2017.

\bibitem{athalye2018obfuscated}
A.~Athalye, N.~Carlini, and D.~Wagner, ``Obfuscated gradients give a false
  sense of security: Circumventing defenses to adversarial examples,'' in
  \emph{International conference on machine learning}.\hskip 1em plus 0.5em
  minus 0.4em\relax PMLR, 2018, pp. 274--283.

\bibitem{goodfellow2014explaining}
I.~J. Goodfellow, J.~Shlens, and C.~Szegedy, ``Explaining and harnessing
  adversarial examples,'' \emph{arXiv preprint arXiv:1412.6572}, 2014.

\bibitem{mandlekar2017adversarially}
A.~Mandlekar, Y.~Zhu, A.~Garg, L.~Fei-Fei, and S.~Savarese, ``Adversarially
  robust policy learning: Active construction of physically-plausible
  perturbations,'' in \emph{2017 IEEE/RSJ International Conference on
  Intelligent Robots and Systems (IROS)}.\hskip 1em plus 0.5em minus
  0.4em\relax IEEE, 2017, pp. 3932--3939.

\bibitem{kuang2022learning}
Y.~Kuang, M.~Lu, J.~Wang, Q.~Zhou, B.~Li, and H.~Li, ``Learning robust policy
  against disturbance in transition dynamics via state-conservative policy
  optimization,'' in \emph{Proceedings of the AAAI Conference on Artificial
  Intelligence}, vol.~36, no.~7, 2022, pp. 7247--7254.

\bibitem{sutton2018reinforcement}
R.~S. Sutton and A.~G. Barto, \emph{Reinforcement learning: An
  introduction}.\hskip 1em plus 0.5em minus 0.4em\relax MIT press, 2018.

\bibitem{haarnoja2018soft}
T.~Haarnoja, A.~Zhou, K.~Hartikainen, G.~Tucker, S.~Ha, J.~Tan, V.~Kumar,
  H.~Zhu, A.~Gupta, P.~Abbeel \emph{et~al.}, ``Soft actor-critic algorithms and
  applications,'' \emph{arXiv preprint arXiv:1812.05905}, 2018.

\bibitem{todorov2012mujoco}
E.~Todorov, T.~Erez, and Y.~Tassa, ``Mujoco: A physics engine for model-based
  control,'' in \emph{2012 IEEE/RSJ international conference on intelligent
  robots and systems}.\hskip 1em plus 0.5em minus 0.4em\relax IEEE, 2012, pp.
  5026--5033.

\end{thebibliography}

\end{document}